\documentclass[a4paper,11pt]{article}

\usepackage{geometry}										
\usepackage[english]{babel}									
\usepackage[utf8]{inputenc}									
\usepackage[T1]{fontenc}									

\usepackage{lmodern}										

\usepackage{amssymb,amsfonts,amsthm,amsmath}
\usepackage[fleqn]{mathtools}
\usepackage{hyphenat}
\usepackage{booktabs, multirow}
\usepackage{bbm}
\usepackage{graphicx}
\usepackage{algpseudocode}
\usepackage{algorithm}
\usepackage{subfigure}
\usepackage{float} 
\usepackage{placeins}

\newcommand{\quot}[1]{``#1''}

\graphicspath{{Figures/}}

\makeatletter
\let\oldabs\abs
\def\abs{\@ifstar{\oldabs}{\oldabs*}}
\let\oldnorm\norm
\def\norm{\@ifstar{\oldnorm}{\oldnorm*}}
\makeatother

\newcommand{\ve}[1]{\boldsymbol{#1}}
\newcommand{\ua}{\ve{\alpha}}
\newcommand{\ca}{{\mathcal A}}
\newcommand{\cd}{{\mathcal D}}
\newcommand{\cm}{{\mathcal M}}

\newcommand{\cs}{{\mathcal S}}
\newcommand{\cx}{{\mathcal X}}
\newcommand{\cy}{{\mathcal Y}}
\newcommand{\Pro}{\mathbb{P}}
\newcommand{\Espe}[2]{{\mathbb E}_{#1}\left[#2\right]}
\newcommand{\GLaM}{\mathrm{GLaM}}
\newcommand{\D}{\mathrm{d}}
\DeclareMathOperator{\DL}{D_{\text{KL}}}

\usepackage{authblk}										

\title{MF-GLaM: A multifidelity stochastic emulator using generalized lambda models}
\author[1]{Katerina Giannoukou \thanks{katerina.giannoukou@ibk.baug.ethz.ch}}
\author[2]{Xujia Zhu \thanks{xujia.zhu@l2s.centralesupelec.fr}}
\author[1]{Stefano Marelli\thanks{marelli@ibk.baug.ethz.ch}}
\author[1]{Bruno Sudret\thanks{sudret@ethz.ch}}
\affil[1]{Chair of Risk, Safety and Uncertainty Quantification, ETH Z\"{u}rich, Switzerland}
\affil[2]{Laboratory of Signals and Systems, CentraleSupélec, Université Paris-Saclay}

\date{\today}

\usepackage{microtype}										

\usepackage{indentfirst}									

\usepackage{caption}
\usepackage{subcaption}

\usepackage{natbib}
\usepackage[hidelinks, colorlinks=true, linkcolor=black, citecolor=black]{hyperref}
\hypersetup{
pdftitle = {MF-GLaM: A multifidelity stochastic emulator using generalized lambda models},
pdfauthor = {Katerina Giannoukou, Xujia Zhu, Stefano Marelli, Bruno Sudret},
pdfsubject = {},
pdfkeywords = {multifidelity, surrogate modeling, stochastic emulators, MF-GLaM, generalized lambda models},
colorlinks = false,                                                                                        
}

\usepackage[capitalize]{cleveref}								
\creflabelformat{equation}{#2(#1)#3}								
\crefrangelabelformat{equation}{#3(#1)#4 to #5(#2)#6}						
\labelcrefformat{subequation}{#2(#1)#3}								
\labelcrefrangeformat{subequation}{#3(#1)#4 to #5(#2)#6}					

\graphicspath{{Figures}} 

\begin{document}

\maketitle

\begin{abstract}
Stochastic simulators exhibit intrinsic stochasticity due to unobservable, uncontrollable, or unmodeled input variables, resulting in random outputs even at fixed input conditions. 
Such simulators are common across various scientific disciplines; however, emulating their entire conditional probability distribution is challenging, as it is a task traditional deterministic surrogate modeling techniques are not designed for.
Additionally, accurately characterizing the response distribution can require prohibitively large datasets, especially for computationally expensive high-fidelity (HF) simulators.
When lower-fidelity (LF) stochastic simulators are available, they can enhance limited HF information within a multifidelity surrogate modeling (MFSM) framework. 
While MFSM techniques are well-established for deterministic settings, constructing multifidelity emulators to predict the full conditional response distribution of stochastic simulators remains a challenge. 
In this paper, we propose multifidelity generalized lambda models (MF-GLaMs) to efficiently emulate the conditional response distribution of HF stochastic simulators by exploiting data from LF stochastic simulators.
Our approach builds upon the generalized lambda model (GLaM), which represents the conditional distribution at each input by a flexible, four-parameter generalized lambda distribution.
MF-GLaMs are non-intrusive, requiring no access to the internal stochasticity of the simulators nor multiple replications of the same input values.
We demonstrate the efficacy of MF-GLaM through synthetic examples of increasing complexity and a realistic earthquake application. 
Results show that MF-GLaMs can achieve improved accuracy at the same cost as single-fidelity GLaMs, or comparable performance at significantly reduced cost.
\end{abstract}

\section{Introduction}

Computer simulations are indispensable tools in engineering and a wide range of applied sciences. 
They predict the behavior of systems based on pre-existing knowledge of the underlying physics. 
This typically involves identifying a set of input variables that affect the system and modeling the interactions between these variables to map them to an output, which represents a quantity of interest.
Most conventional simulators are \emph{deterministic}, meaning that for a given set of input parameters $\ve{x}$, they always produce the same response $y = \cm_d (\ve{x}) \in \mathbb{R}$. 
For example, finite-element simulators used in structural engineering provide deterministic predictions of stress and strain distributions under specific load conditions.

However, deterministic simulators are sometimes inadequate for modeling complex systems.
For example, in many experimental settings, some input variables that influence the output are not directly observable or controllable.
A common example is the microstructural features of a material in material testing, which are often unknown but could significantly affect its mechanical behavior.
In other cases, models exhibit intrinsic stochasticity, such as those commonly found in economics or epidemiology, e.g., stock price evolution or disease spread models.
Moreover, when the inputs are very high-dimensional, they are often reduced to summary statistics, as seen in wind turbine simulations or earthquake engineering, thus introducing latent variability in outputs. 

Such systems are modeled using \emph{stochastic simulators}, where the unobservable, uncontrollable, or unmodeled inputs can be considered as latent variables and introduce inherent stochasticity into the system response. 
Then, the model response $\cm_s (x)$ at any fixed input $\ve{x}$ is a random variable. Repeated model evaluations at the same $\ve{x}$ generate different realizations of this random variable, and they are referred to as \emph{replications}.
Formally, a stochastic simulator is defined by
\begin{equation}
\begin{split}
    \mathcal{M}_s : \mathcal{D}_{\ve X} \times \Omega &\to \mathbb{R}, \\
    (\ve x, \omega) &\mapsto \mathcal{M}_s(\ve x, \omega),
\end{split}
\end{equation}
where $\ve x$ is the input vector in the input space $\mathcal{D}_{\ve X}$, and $\Omega$ denotes the probability space which represents the internal stochasticity.
For stochastic simulators, the statistical properties of the response conditioned on $\ve{x}$, such as the mean, quantiles, or the entire conditional distribution are often of primary interest.

In contrast, our goal is to obtain the \emph{entire} conditional distribution produced by a stochastic simulator. 
In principle, this requires numerous replications at each input $\ve{x}$ to fully characterize the corresponding conditional response distribution, making the task intractable for computationally expensive computer models.
Hence, we wish to develop \emph{surrogate models}, also called \emph{emulators}, that accurately predict the entire probability distribution function (PDF) of the simulator’s output at any given input at a fraction of the original simulation cost.

While emulating deterministic simulators is a mature field of research, commonly employing techniques such as polynomial chaos expansions (PCE) \citep{Xiu2002,BlatmanJCP2011}, Gaussian processes (GPs) \citep{Rasmussen2006}, and support vector regression \citep{Drucker1996}, emulating stochastic simulators has only recently begun to receive comparable attention.
According to \citet{LuethenCMAME2023}, there are two main approaches for emulating stochastic simulators, each reflecting a different perspective. The first, called the random function view, treats the stochastic simulator output as a random field over the input space. 
Such methods \citep{AzziIJUQ2019,LuethenCMAME2023,Mueller2025} are able to reproduce the dependence structure of the responses at different input parameters, by emulating different trajectories, each of which is a deterministic function of the input $\ve x$ for fixed internal stochasticity $\omega$ of the stochastic simulator.
However, in many cases, access to the internal stochasticity is unavailable, or the focus is solely on emulating the conditional response distribution at each input $\ve x$. 
Then, the second class of methods, which adopt a random variable view, becomes pertinent. 
Within this class, one category relies on replications at different input locations to characterize the simulator's response distributions by isolating its stochasticity. 
Replication-based approaches include stochastic Kriging \citep{Ankenman2010}, which focuses on estimating summary statistics; nonparametric methods such as those proposed by \citet{Moutoussamy2015,Browne2016}, which aim to estimate the full response PDF; and the parametric framework introduced by \citet{ZhuIJUQ2020}.
A second category requires neither replications nor access to the internal stochasticity.
This includes both nonparametric methods, such as kernel density estimators \citep{Hall2004}, and parametric methods, including the generalized lambda model approach \citep{ZhuSIAM2021} and the stochastic PCE approach \citep{ZhuStoPCE2023}.

Emulating stochastic simulators can be computationally demanding due to the large number of model evaluations required to estimate full response distributions accurately. When emulating expensive high-fidelity (HF) models,  the associated computational cost can be prohibitive. 
In scenarios where stochastic simulators of different fidelities exist, we can take advantage of the lower-fidelity (LF) model to emulate the higher fidelity one. 
This is the core idea behind \emph{multifidelity surrogate modeling}.
Multifidelity surrogate models (MFSMs) combine data from multiple sources of different fidelity into a single surrogate model, typically enhancing a limited, computationally expensive HF dataset with more extensive and less expensive LF ones.
Over the past two decades, multifidelity surrogate modeling has been extensively studied in the context of deterministic simulators, leading to a variety of successful approaches, including Gaussian process (GP)-based methods \citep{kennedy2000,le2014recursive}, deep GP-based approaches \citep{Cutajar2018,Hebbal2021}, PCE-based approaches \citep{Ng2012,Palar2016}, and neural network-based frameworks \citep{Zhang2022,Conti2023}.
However, not all multifidelity approaches build a single combined surrogate model. In probabilistic UQ, for example, fidelity hierarchies can be directly exploited via Monte Carlo methods, such as multilevel Monte Carlo \citep{Giles2008}, approximate control‐variate estimators \citep{Gorodetsky2020}, multifidelity importance sampling \citep{Dubourg2014,Peherstorfer2016}, and control variates importance sampling \citep{Chakroborty2024}. 
Likewise, in interval‐based UQ, multilevel interval‐analysis methods have been developed \citep{Callens2022}.

Moreover, some methods also address \emph{noisy} models, where the system itself is considered deterministic, but the observed outputs contain external noise, such as measurement or observational noise, which is typically treated as homoscedastic.
The primary objective in these settings is to denoise the data and recover the underlying mean system response using regression-based approaches, such as linear regression \citep{Zhang2018}, PCE \citep{GiannoukouASCE2025}, GP-regression \citep{Forrester2007}, and neural networks \citep{Meng2021}. 

However, constructing multifidelity surrogate models to emulate inherently stochastic simulators, where variability arises intrinsically and the homoscedasticity assumption is relaxed, has received comparatively little attention until recently. 
\citet{Perdikaris2015} propose a recursive co-Kriging framework to integrate multifidelity information both in the models and in the probability space for design under uncertainty, focusing on estimating the mean or other statistical quantities of interest of a HF simulator. 
The first attempt to apply multifidelity modeling to correlated stochastic processes for learning the conditional distribution of the HF process is proposed by \citet{Yang2019a}, who  propose a surrogate model by performing variational inference using adversarial learning.
Moreover, \citet{Stroh2021} introduce an adaptive sampling scheme based on GP modeling, applicable to both deterministic and stochastic multifidelity models. 
In addition, \citet{Bae2020} develop a nondeterministic localized‑Galerkin approach extending the nondeterministic Kriging \citep{Bae2019a} that can handle both aleatoric and epistemic uncertainties.
\citet{Tao2025} propose an MF co-Kriging approach for stochastic simulators with heteroscedastic noise, while \citet{Konomi2023} propose a Bayesian latent‑variable co‑Kriging model for handling quality‑flagged observations in remote sensing applications.  
In the relevant context of uncertainty quantification, \citet{Reuter2024} extend the approximate control variate framework \citep{Gorodetsky2020} for multifidelity UQ in stochastic simulators, yet without constructing surrogate models that emulate the entire output distribution.
Although these methods show significant progress, certain challenges remain. 
Several approaches focus on specific summary statistics, such as the mean, variance, or selected quantiles, rather than modeling the full conditional response distribution of high-fidelity stochastic emulators.
In addition, a generally applicable and computationally efficient multifidelity stochastic emulator that can flexibly capture highly non-Gaussian response distributions from limited stochastic data has yet to be established.

To address the gap, we propose a novel multifidelity framework for emulating the \emph{entire} conditional response distribution of high-fidelity stochastic simulators. 
Our approach extends the single-fidelity generalized lambda model (GLaM) introduced by \citet{ZhuSIAM2021}, in which the response PDF of a stochastic simulator is approximated by the flexible parametric family of generalized lambda distributions (GLDs). 
To extend this parametric method to the multifidelity setting, we propose to fuse the variable-fidelity information directly at the level of the GLD parameters, expressing each high-fidelity parameter as its low-fidelity counterpart plus a learned discrepancy.
The proposed method is non-intrusive, as it does not require knowledge of the simulator’s internal workings, thus treating the HF and LF simulators as \quot{black boxes}.
Moreover, our method does not require replications at the same HF or LF inputs; instead, each model provides a single random output per input sample.
These data are then combined into a multifidelity stochastic emulator, referred to as the \emph{multifidelity generalized lambda model} (MF-GLaM), that enables inexpensive, full-distribution predictions of high-fidelity stochastic simulator responses.
We validate our methodology on three increasingly complex examples, comprising two synthetic models and a real-world earthquake engineering application. 

The remainder of this paper is organized as follows. 
Firstly, we recall the GLaM methodology in the single-fidelity setting in \cref{sec:GLaM_single}, along with the relevant theory.
Next, we introduce the MF-GLaM framework and evaluate its performance on three numerical examples, in \cref{sec:GLaM_MF,sec:applications}, respectively. 
Finally, in \cref{sec:conclusions}, we draw conclusions and discuss prospects for future research.


\section{Generalized lambda models (GLaMs) in single fidelity}
\label{sec:GLaM_single}
The generalized lambda model framework, originally introduced by \citet{ZhuIJUQ2020,ZhuSIAM2021}, provides a surrogate that emulates the entire conditional response distribution of a stochastic simulator at any given input $\ve{x}$. 
Given a stochastic simulator $\mathcal{M}_s$, its response $\mathcal{M}_s(\ve{x}, \omega)$ at a fixed input $\ve{x}$ is represented by the random variable $Y_{\ve{x}}$. 
Let us consider an experimental design $(\ve{\cx}, \cy)$, where $\ve{\cx} = \{\ve{x}^{(1)}, \dots, \ve{x}^{(N)}\}$ are the input samples, and $\cy = \{\mathcal{M}_s(\ve{x}^{(1)}, \omega^{(1)}), \dots, \mathcal{M}_s(\ve{x}^{(N)}, \omega^{(N)})\}$ are the corresponding outputs. 
Please note that no common random numbers are enforced; the realizations $\omega^{(1)}, \dots,\omega^{(N)}$ may differ from one design point to another.
For notational simplicity, we write $\cy = \{y^{(1)}, \dots, y^{(N)}\}$. 
The core assumption in the GLaM framework is that, for each input $\ve{x}$, the random variable $Y_{\ve{x}}$ can be accurately approximated by the family of generalized lambda distributions:
\begin{equation}
     Y_{\ve x} \sim \operatorname{GLD}\left(\lambda_1(\ve{x}), \lambda_2(\ve{x}), \lambda_3(\ve{x}), \lambda_4(\ve{x})\right),
     \label{eq:y_sim_GLD}
\end{equation}
where  $\ve \lambda = \{\lambda_1, \lambda_2, \lambda_3, \lambda_4\}$ are the four distribution parameters.

The GLD is a highly flexible distribution family capable of approximating most common unimodal distributions, such as uniform, Gaussian, Student’s t, exponential, lognormal, and Weibull distributions.
The GLD parametrizes the quantile function, i.e., the inverse of the cumulative distribution function, defined as follows:
\begin{equation}
\label{eq:GLD quantile}
    Q(u; \ve{\lambda}) = \lambda_1 + \frac{1}{\lambda_2} \left( \frac{u^{\lambda_3} - 1}{\lambda_3} - \frac{(1 - u)^{\lambda_4} - 1}{\lambda_4} \right).
\end{equation}
The PDF of a random variable \( Y\) following a GLD can be computed as follows:
\begin{equation}
\label{eq:pdf_GLD}
    f_Y^{\text{GLD}}(y) = \frac{\lambda_2}{u^{\lambda_3 - 1} + (1 - u)^{\lambda_4 - 1}} \mathbbm{1}_{[0,1]}(u), \quad \text{with } u = Q^{-1}(y),
\end{equation}
where $\mathbbm{1}_{[0,1]}$ is the indicator function. A closed-form expression for \( Q^{-1} \), and consequently for \( f_Y \), is generally not available.

As for the roles of the four parameters, $\lambda_1$ acts as a location parameter, $\lambda_2$ as a scale parameter, and $\lambda_3,\lambda_4$ as shape parameters. To guarantee a valid quantile function, $\lambda_2$ must be strictly positive. 
Furthermore, the values of $\lambda_3$ and $\lambda_4$ determine the shape and boundedness of the distribution. For instance, if $\lambda_3 = \lambda_4$, the resulting distribution is symmetric, while $\lambda_3 > 0$ or $\lambda_4 > 0$ produce left- or right-bounded distributions, respectively.

For $\lambda_3, \lambda_4 > -0.5$, the mean and variance of $Y$ exist and are given by:
\begin{align}
    &\mathbb{E}[Y] = \lambda_1 - \frac{1}{\lambda_2} \label{eq:GLD_mean}
    \left( \frac{1}{\lambda_3 + 1} - \frac{1}{\lambda_4 + 1} \right),\\
    &\text{Var}[Y] = \frac{d_2 - d_1^2}{\lambda_2^2}, \label{eq:GLD_var}
\end{align}
where $d_1$ and $d_2$ are defined as follows:
\begin{align}
    d_1 &= \frac{1}{\lambda_3} B(\lambda_3 + 1, 1) - \frac{1}{\lambda_4} B(1, \lambda_4 + 1), \\
    d_2 &= \frac{1}{\lambda_3^2} B(2\lambda_3 + 1, 1) - \frac{2}{\lambda_3 \lambda_4} B(\lambda_3 + 1, \lambda_4 + 1) 
    + \frac{1}{\lambda_4^2} B(1, 2\lambda_4 + 1),
\end{align}
where $B(a, b)$ denotes the Beta function.
For further details on the properties of GLDs, the reader is referred to \citet{Freimer1988,ZhuSIAM2021}.

When applied to stochastic emulation, each of the four parameters $\ve{\lambda}$ in \cref{eq:y_sim_GLD} is modeled as a function of $\ve{x}$, allowing the shape of the distribution to vary throughout the input space, a property necessary to address heteroskedasticity in the model response.
To achieve this, \citet{ZhuSIAM2021} construct a functional approximation of each component of $\ve{\lambda}(\ve x)$ using polynomial chaos expansions.
%
The latter is a surrogate modeling technique that approximates the response of a (deterministic) model with finite variance through its spectral representation on an orthogonal polynomial basis \citep{Xiu2002,ghanem_spanos_1991}.
In practice, this series is truncated to a finite number of terms.
Consider a random vector $\ve{X}\in \mathbb{R}^M $ with independent components and joint PDF $f_{\ve{X}}(\ve x) = \prod_{i=1}^{M}f_{{X}_i}(x_i),$ where $f_{{X}_i}$ is the marginal PDF of the random variable ${X}_i$.
The truncated PCE of a model $\cm(\ve{x})$ is expressed as
\begin{equation}
\label{eq:PCE_trunc}
    \widetilde{\mathcal{M}}^\text{PC} \left( \ve{x} \right) = \sum_{\ve\alpha \in \ca} c_{\ve\alpha} {\Psi}_{\ve\alpha} \left( \ve{x} \right),
\end{equation}
where $c_{\ua} \in \mathbb{R}$ are the expansion coefficients, and \{$\Psi_{\ua}, \,\ua \in \ca$\} are the multivariate polynomials forming the basis. 
Each polynomial $\Psi_{\ua}$, characterized by the multi-index $\ua$, is the product of univariate polynomials, orthogonal with respect to $f_{\ve{X}_i}$.
The finite set of multi-indices $\ca \subset \mathbb{N}^M$ determines which polynomials are included in the expansion, and can be obtained from various truncation schemes \citep{BlatmanJCP2011,LuethenSIAMJUQ2021}.
Hyperbolic, a.k.a. q-norm truncation is a common choice, resulting in a set $\ca_{p,q,M}$, given by:
\begin{equation}
    \ca_{p,q,M} = \left\{ \boldsymbol{\alpha} \in \mathbb{N}^M, \|\boldsymbol{\alpha}\|_q = \left( \sum_{i=1}^{M} |\alpha_i|^q \right)^{\frac{1}{q}} \leq p \right\},
\end{equation}
where $p$ is the maximum degree of polynomials, $0 < q \leq 1$ defines the quasi-norm $\|\cdot\|_q $, with $q = 1$ resulting in the full basis of total degree less than $p$.

Returning to the four parameters of the GLD, each can be approximated by PCE as follows:
\begin{align}
    &\lambda_i (\ve x) \approx \lambda_i^{\text{PC}} (\ve x; \ve{c}_i) = \sum_{\ua \in \ca_i} c_{i,\ua} \psi_{\ua}(\ve x), \quad i = 1,3,4, \\
    &\lambda_2 (\ve x) \approx \lambda_2^{\text{PC}} (\ve x; \ve{c}_2) = \exp \left( \sum_{\ua \in \ca_2} c_{2,\ua} \psi_{\ua}(\ve x) \right).
\end{align}
To ensure the positiveness of $\lambda_2$, its PCE representation is constructed using an exponential transform.

Constructing a generalized lambda model reduces to selecting appropriate truncation sets $\ve \ca = \{ \mathcal{A}_l : l = 1, \dots, 4 \}$ and estimating the corresponding PCE coefficients $\ve c$ based on the available data $(\ve \cx, \cy)$.

For given truncation sets $\ve{\mathcal{A}}$, the coefficient vector $\ve{c}$ can be estimated using maximum likelihood estimation (MLE):
\begin{equation}
\label{eq:max_likel}
    \ve{c}^* = \arg \max_{\ve{c}} \log L(\ve{c}; \ve{\cx}, \cy),
\end{equation}

where $\log L(\ve{c}; \ve{\cx}, \cy)$ is the log-likelihood function, given by:
\begin{equation}
    \log L(\ve{c}; \ve{\cx}, \cy) = \sum_{i=1}^{N} \log \left( f^{\text{GLD}} \left( y^{(i)}; \lambda^{\text{PC}} (\ve x^{(i)}; \ve{c}) \right) \right),
\end{equation}
where $f^{\text{GLD}}$ is the PDF of the GLD, as defined in \cref{eq:pdf_GLD}.

The truncation sets $\ca_1$ and $\ca_2$ corresponding to $\lambda_1^{\text{PC}}$ and $\lambda_2^{\text{PC}}$, respectively, are obtained using a modified feasible generalized least-squares approach \citep{ZhuSIAM2021}. 
More precisely, the mean and variance functions of the model response are each expressed as PCEs and are fitted alternatively and iteratively using the sparse solver hybrid least-angle regression \citep{BlatmanJCP2011}. 
These fitted PCEs provide the truncation sets for $\lambda_1^{\text{PC}}$ and $\lambda_2^{\text{PC}}$, as well as starting points $\ve c_{1,0}, \ve c_{2,0}$ for the optimization in \cref{eq:max_likel}.

To determine appropriate truncation sets $\ca_3$ and $\ca_4$ for $\lambda_3^{\text{PC}}$ and $\lambda_4^{\text{PC}}$, the Bayesian information criterion (BIC; \citet{Schwartz1978}) can be employed for model selection. This is defined as:
\begin{equation}
    \text{BIC} \coloneqq -2\log L(\ve{c}^*; \ve{\cx}, \cy) + \log(N) \|\ve{c}^*\|_0,
\end{equation}
where $L$ is the likelihood function, $\ve{c}^*$ denotes the estimated coefficient vector, and $\|\ve{c}^*\|_0$ is the $\ell_0$-norm corresponding to the number of nonzero elements in $\ve{c}^*$.
Lower BIC values indicate models with a better trade-off between goodness of fit and complexity.

To select $\ca_3$ and $\ca_4$, different GLaMs are constructed with increasing polynomial degrees of $\lambda_3^{\text{PC}}$ and $\lambda_4^{\text{PC}}$ and the BIC is computed at each step.  
The truncation sets $\ca_3$ and $\ca_4$ are chosen as those beyond which further increases in polynomial degree yield no improvement in the BIC.

For further details on the GLaM construction, the reader is referred to \citet{ZhuSIAM2021,ZhuPEM2023} and the corresponding implementation in the UQLab software \citep{Marelli2014,UQdoc_21_120_GLaM}.

\section{Multifidelity generalized lambda models (MF-GLaMs)}
\label{sec:GLaM_MF}

\subsection{Introduction}
In this section, we provide the conceptual and methodological details for constructing multifidelity stochastic emulators based on generalized lambda models.

We begin by considering an expensive-to-evaluate high-fidelity stochastic simulator, denoted as $\mathcal{M}_{s}^{\text{H}}(\ve x_\text{H}, \omega_\text{H})$, where $\ve x_\text{H}$ is a realization of the random input vector $\ve X_\text H$, and $\omega_\text{H} \in \Omega_\text{H}$ is an abstract random event representing the internal stochasticity.
Our goal is to emulate the conditional response distribution $Y_\text{H} \mid \ve X_\text{H} = \ve x_\text{H}$ of $\mathcal{M}_{s}^{\text{H}}$ for each $\ve x_\text{H}$, which we denote for convenience as $Y_{\ve{x}_\text{H}}^\text{H}$.

Suppose that, in addition, a lower-fidelity, cheaper-to-evaluate stochastic simulator $\mathcal{M}_{s}^{\text L}(\ve x_\text L, \omega_\text L)$ is available, with $\ve x_\text L$ being a realization of the input random vector $\ve X_\text L$, and $\omega_\text L \in \Omega_\text L$ representing the internal stochasticity in this LF stochastic simulator. 
We similarly denote the conditional LF response distribution  $Y_\text L \mid \ve X_\text L = \ve x_\text L$ of the LF simulator as $Y_{\ve x_\text L}^\text L$.

Moreover, we assume that the input variables of the LF model form a subset of those of the HF model, i.e. $\ve x_\text L \subseteq \ve x_\text H$. 
To simplify the notation, we hereafter denote both the HF and LF inputs $\ve x_\text H,\, \ve x_\text L$ simply by $\ve{x}$, and the associated random vectors by  $\ve X$.
When $\ve x_\text L \subset \ve x_\text H$, the low-fidelity simulator ignores the components of $\ve x$ it does not use.
Without loss of generality, we focus on the case where exactly two levels of fidelity are present.

A multifidelity stochastic emulator aims to approximate the conditional response distribution $Y_{\ve{x}}^\text{H}$ of the HF stochastic simulator by combining information from all the available variable-fidelity models. 
Here, we propose to construct such an emulator based on the GLaM stochastic emulator methodology.
To this end, we assume that the response distributions of both stochastic simulators follow a generalized lambda distribution:
\begin{equation} 
    Y_{\ve x}^\text H \sim \operatorname{GLD}\left(\lambda_1^\text H(\ve{x}), \lambda_2^\text H(\ve{x}), \lambda_3^\text H(\ve{x}), \lambda_4^\text H(\ve{x})\right) 
\end{equation}  
and
\begin{equation} 
    Y_{\ve x}^\text L \sim \operatorname{GLD}\left(\lambda_1^\text L(\ve{x}), \lambda_2^\text L(\ve{x}), \lambda_3^\text L(\ve{x}), \lambda_4^\text L(\ve{x})\right).
\end{equation}

Then, the MF stochastic emulator denoted as $\tilde Y_{\ve x}^\text{MF}$, also follows a GLD:
\begin{equation} 
    Y_{\ve x}^\text H \stackrel{\mathrm{d}}{\approx} \tilde Y_{\ve x}^\text{MF}  \sim \operatorname{GLD}\left(\lambda_1^\text{MF}(\ve{x}), \lambda_2^\text{MF}(\ve{x}), \lambda_3^\text{MF}(\ve{x}), \lambda_4^\text{MF}(\ve{x})\right), 
\end{equation}  
where $\stackrel{\mathrm{d}}{\approx}$ denotes approximate equality in distribution.
We refer to this multifidelity emulator as the multifidelity generalized lambda model (MF-GLaM).

The parameters $\ve{\lambda}^\text{MF}$ aim to approximate the parameters $\ve{\lambda}^\text{H}$ of the HF GLD.
Assuming that the discrepancy between the HF and LF GLaM parameters, $\ve \lambda^\text H$ and $\ve \lambda^\text L$, is less complex to model than the HF parameters themselves, we propose to express each  MF-GLaM parameter as the sum of the corresponding LF parameter and a discrepancy function:
\begin{equation} \label{eq:lambda_MF}
    \lambda_i^\text{MF}(\ve{x}) = \lambda_i^\text L(\ve{x}) + \delta_i (\ve x), \quad i = 1, ...,4 .
\end{equation}
This formulation is directly inspired by the deterministic multifidelity fusion scheme introduced by \citet{kennedy2000} and subsequently employed with PCE as a surrogate by \citet{Ng2012}, here adapted to the stochastic context.

In our approach, we similarly use PCE to represent the LF GLaM parameters:
\begin{align}
    &\lambda_i^\text{L} (\ve x) \approx 
    \lambda_i^{\text{L, PC}} (\ve x; \ve{c}) = 
    \sum_{\ua \in \ca_i^\text L} c_{i,\ua} \psi_{\ua}(\ve x), \quad i = 1,3,4, \label{eq:lambda_134_LF}\\
    &\lambda_2^\text{L} (\ve x) \approx 
    \exp \left( \lambda_2^{\text{L, PC}} (\ve x; \ve{c}) \right ) = 
    \exp \left( \sum_{\ua \in \ca_2^\text L} c_{2,\ua} \psi_{\ua}(\ve x) \right ). \label{eq:lambda_2_LF}
\end{align}

We also use PCE to model the discrepancy functions $\ve{\delta}$ in \cref{eq:lambda_MF}. Hence, $\ve \lambda^\text{MF}$ is written as:
\begin{align}
    &\lambda_i^\text{MF} (\ve x) \approx 
    \lambda_i^{\text{L, PC}} (\ve x; \ve{c}) + \delta_i^{\text{PC}} (\ve x; \ve{d}) = 
    \sum_{\ua \in \ca_i^\text L} c_{i,\ua} \psi_{\ua}(\ve x) + \sum_{\ua \in \ca_i^\delta} d_{i,\ua} \psi_{\ua}(\ve x), \quad i = 1,3,4, \label{eq:lambda_134_MF}\\
    &\lambda_2^\text{MF} (\ve x) \approx 
    \exp \left( \lambda_2^{\text{L, PC}} (\ve x; \ve{c}) + \delta_2^{\text{PC}} (\ve x; \ve{d}) \right ) = 
    \exp \left( \sum_{\ua \in \ca_2^\text L} c_{2,\ua} \psi_{\ua}(\ve x) + \sum_{\ua \in \ca_2^\delta} d_{2,\ua} \psi_{\ua}(\ve x) \right ), \label{eq:lambda_2_MF}
\end{align}
where the exponential transform for $\lambda_2^\text{MF}(\ve{x})$ ensures it remains positive.
Here,\\ $\ve \ca^\text L = \{\ca_i^\text L: i=1,...,4\}$ defines the truncation sets specifying the PCE basis for the LF parameters, with associated coefficients $\ve c = \{c_{i,\ve \ua}: i=1,...,4, \, \ua \in \ca_i^\text L\}$. 
Similarly, \\
$\ve{\ca}^\delta = \{\ca_i^\delta: i=1,...,4\}$ denotes the truncation sets for the discrepancy expansions, with associated coefficients $\ve d = \{d_{i,\ve \ua}: i=1,...,4, \, \ua \in \ca_i^\delta\}$. 

Hence, constructing the MF-GLaM for emulating $Y_{\ve{x}}^\text{H}$ involves:
\begin{enumerate}
\item defining the sets $\ve \ca^\text{L}$ and $\ve \ca^\delta$, and 
\item estimating the full set of unknown coefficients, denoted as $\ve \theta$:
  \begin{equation}
    \ve \theta = (\ve c, \ve d).
  \end{equation}
\end{enumerate}
In the following sections, we describe the procedure for accomplishing these steps.

\subsection{Estimation of the model parameters}
\label{sec:estimation}
Consider a high-fidelity dataset $(\ve \cx_\text H, \cy_\text H)$, where 
$\ve \cx_\text H = \{ \ve x_\text H ^{(i)}: \, i = 1, ..., N_\text H\}$ and  \\
$\cy_\text H = \{ y_\text H ^{(i)} = \cm_\text H (\ve x_\text H ^{(i)}, \omega_\text H^{(i)}): \, i = 1, ..., N_\text H\}$, 
along with a low-fidelity dataset $(\ve \cx_\text L, \cy_\text L)$, where
$\ve \cx_\text L = \{ \ve x_\text L ^{(i)}: \, i = 1, ..., N_\text L\}$ and  
$\cy_\text L = \{ y_\text L ^{(i)} = \cm_\text L (\ve x_\text L ^{(i)}, \omega_\text L^{(i)}): \, i = 1, ..., N_\text L\}$. 
It is reminded that both HF and LF stochastic simulators are evaluated only once per input sample, as no replications are required.
In this section, we derive the maximum likelihood estimator for the unknown MF-GLaM coefficients $\ve\theta$ from the available HF and LF data, given the truncation sets $\ve \ca^\text L$ and $\ve \ca^\delta$. The procedure to obtain these truncation sets, as well as the fitting procedure, are detailed in the following section.

From \cref{eq:lambda_134_LF,eq:lambda_2_LF,eq:lambda_134_MF,eq:lambda_2_MF}, we observe that some of the coefficients in $\ve\theta$ appear in both the LF GLaM and the MF-GLaM, namely, the coefficients $\ve{c}$.
On the other hand, the coefficients $\ve{d}$ appear only in the MF-GLaM.
Our goal is thus to jointly estimate \emph{all} unknown coefficients $\ve{\theta}$ using both the HF and LF data simultaneously. 
To achieve this, we employ a maximum likelihood estimation approach.
More precisely, we derive such an estimator by minimizing the Kullback-Leibler divergence between the true underlying joint distribution of the HF and LF inputs and outputs, and the distribution approximated by GLaMs.
To do this, we will need to delve deeper into the nature of multifidelity modeling for stochastic models.

In multifidelity analysis, we typically observe data of the form $(s,\ve{x}_s,y_s)$, where $s\in\{L,H\}$ indicates which model is evaluated (low- or high-fidelity), $\ve{x}_s$ denotes the model's input, and $y_s$ is the corresponding output. 
%
We define the following sets:
\begin{equation}
\mathcal{D}_S = \{L,H\}, 
\quad
\mathcal{D}_{\ve{X}} \subseteq \mathbb{R}^n,
\quad
\mathcal{D}_{Y_H} \subseteq \mathbb{R},
\quad
\mathcal{D}_{Y_L} \subseteq \mathbb{R}.
\end{equation}

Thus, the random variables $S,\ve{X},Y_H,Y_L$ take values in $\mathcal{D}_S,\mathcal{D}_{\ve{X}}, \mathcal{D}_{Y_H},\mathcal{D}_{Y_L}$, respectively. 
We endow these spaces with the counting measure $\mu_S$ for the discrete label $S$, and Lebesgue measures $\mu_{\ve{X}}, \mu_{Y_H}, \mu_{Y_L}$ on the respective continuous spaces. 
This yields the product measure 
\begin{equation}
\mu_{S,\ve{X},Y_H,Y_L} \;=\; \mu_S \,\otimes\, \mu_{\ve{X}} \,\otimes\, \mu_{Y_H} \,\otimes\, \mu_{Y_L}
\end{equation}
on the product space $\mathcal{D}_S \times \mathcal{D}_{\ve{X}} \times \mathcal{D}_{Y_H} \times \mathcal{D}_{Y_L}$.

Under this setup, the joint probability density function with respect to $\mu_{S,\ve{X},Y_H,Y_L}$ is given by
\begin{equation}
\label{eq:joint_model}
f_0\bigl(s,\ve{x},y_H,y_L\bigr)
\;=\;
p \,\mathbbm{1}_{\{s=L\}} \, f_{\ve{X}}(\ve{x}) \, f_{Y_L \mid \ve X}\bigl(y_L \mid \ve{x}\bigr)
\;+\;
(1-p)\,\mathbbm{1}_{\{s=H\}} \, f_{\ve{X}}(\ve{x}) \, f_{Y_H \mid \ve X}\bigl(y_H \mid \ve{x}\bigr),
\end{equation}
where $\mathbb{P}(S=L) = p$, $\mathbb{P}(S=H) = 1-p$, and $f_{\ve{X}}$ is the PDF of $\ve{X}$. 
Here, $p$ represents the prior probability that the label $S$ takes the value $L$, corresponding to selecting the LF model, and captures the relative weight of sampling LF data compared to the HF data.
Moreover, $f_{Y_L \mid \ve X}\bigl(y_L \mid \ve{x}\bigr)$ and $f_{Y_H \mid \ve X}\bigl(y_H \mid \ve{x}\bigr)$ denote the conditional densities of $Y_L$ and $Y_H$, respectively, given their input variables.
We note that each conditional density depends only on the subset of inputs that the specific model uses. Therefore, in \cref{eq:joint_model} we have ignored the irrelevant input variables when expressing the conditional distributions, e.g., $f_{Y_L \mid \ve X}(y_L\mid \ve{x})$.

A GLaM surrogate for the LF response provides an approximation $f_{Y_L \mid \ve X, \GLaM}\bigl(y_L \mid \ve \lambda^{\text{L, PC}} (\ve x; \ve{c}) \bigr)$ to the conditional PDF $f_{Y_L \mid \ve X}\bigl(y_L \mid \ve{x}\bigr)$. 
For brevity, the GLaM approximation to the LF conditional PDF is denoted as $f_{\GLaM}^\text L\bigl(y_L \mid \ve x; \ve{c} \bigr)$.
Moreover, the multifidelity  GLaM surrogate provides an approximation $f_{Y_H \mid \ve X, \GLaM}\bigl(y_H \mid \ve \lambda^{\text{H, PC}} (\ve x; \ve{c}, \ve{d}) \bigr)$, denoted as $f_{\GLaM}^\text{MF}\bigl(y_H \mid \ve x; \ve{c}, \ve{d} \bigr)$, to the HF conditional PDF $f_{Y_H \mid \ve X}\bigl(y_H \mid \ve{x}\bigr)$.
Thus, the GLaM approximation to the joint PDF in \cref{eq:joint_model} is obtained as follows:
\begin{equation}
\label{eq:joint_model_glam}
    f_{\GLaM}(s,\ve{x},y_H,y_L;\ve{\theta}) = p \mathbbm{1}_{\{s=L\}}f_{\ve{X}}(\ve{x}) f_{\GLaM}^\text{L}(y_L|\ve{x};\ve{c}) + (1-p)\mathbbm{1}_{\{s=H\}}f_{\ve{X}}(\ve{x}) f_{\GLaM}^\text{MF}(y_H|\ve{x};\ve{c}, \ve{d}).
\end{equation}

The log-likelihood estimator to maximize for obtaining the coefficients $\ve{\theta}$ is derived by minimizing the  Kullback--Leibler divergence between the joint distributions $f_0$ and $f_{\GLaM}$, and is given by:
\begin{equation}
\begin{split}
\label{eq:ll_with_p_no_const}
\ell(\ve{\theta};\cs, \ve \cx_\text H, \ve \cx_\text L,\cy_\text H, \cy_\text L) = 
&\frac{p(N_\text{L}+N_\text{H})}{N_\text{L}}\sum_{i=1}^{N_\text{L}} \log\left(f_{\GLaM}^\text{L}(y^{(i)}_\text{L}|\ve{x}_\text{L}^{(i)};\ve{c})\right) \\
&+ \frac{(1-p)(N_\text{L}+N_\text{H})}{N_\text{H}}\sum_{i=1}^{N_\text{H}} \log\left(f_{\GLaM}^\text{MF}(y^{(i)}_\text{H}|\ve{x}_\text{H}^{(i)}; \ve{c}, \ve{d})\right).
\end{split}
\end{equation}
The detailed derivation to obtain this estimator is provided in \cref{annex_A}.

In common multifidelity settings, the HF data are typically available in much smaller quantities. Relying solely on a limited HF dataset (i.e., $p=0$) to train a model could lead to overfitting, since the model does not have enough information to generalize reliably. Thus, the LF data can act as a form of ``regularization''.
A practical compromise, in the absence of domain-specific guidance, is to assign a neutral weight $p=0.5$ between LF and HF in the likelihood function, ensuring that neither fidelity dominates simply because of its dataset size. Under this assumption, from \cref{eq:ll_with_p_no_const}, the log-likelihood to be maximized with respect to the parameters $\ve{\theta}$ is given by:
\begin{equation}\label{eq:llh}
\begin{split}
    \ell(\ve{\theta};\cs, \ve \cx_\text H, \ve \cx_\text L,\cy_\text H, \cy_\text L)  = &\frac{(N_\text{L}+N_\text{H})}{2 N_\text{L}}\sum_{i=1}^{N_\text{L}} \log\left(f_{\GLaM}^\text{L}(y^{(i)}_\text{L}|\ve{x}_\text{L}^{(i)};\ve{c})\right) \\
    &+ \frac{(N_\text{L}+N_\text{H})}{2N_\text{H}}\sum_{i=1}^{N_\text{H}} \log\left(f_{\GLaM}^\text{MF}(y^{(i)}_\text{H}|\ve{x}_\text{H}^{(i)}; \ve{c}, \ve{d})\right).
\end{split}
\end{equation}

Then, the optimal set of PCE coefficients $\ve{\theta}^* = (\ve{c}^*, \ve{d}^*)$ is obtained by:
\begin{equation}
\label{eq:theta_optim}
    \ve{\theta}^*=\arg\max_{\ve{\theta}} \, \ell(\ve{\theta};\cs, \ve \cx_\text H, \ve \cx_\text L,\cy_\text H, \cy_\text L) .
\end{equation}

In some scenarios, one may wish to treat $p$ as a tunable \emph{hyperparameter} to balance the bias-variance trade-off. 
For example, if an application provides more extensive HF data, the HF terms might be given higher weight. Conversely, if HF data are extremely sparse and LF data are sufficiently correlated with the HF response, one could increase the LF weight to better regularize the overall fit. 
Then, $p$ can be learned from the data itself in an adaptive way.


\subsection{Basis selection and fitting procedure}
In this section, we present a method for determining suitable truncation schemes $\ve \ca^\text L, \ve \ca^\delta$ for the expansions of $\{\lambda_i^\text{MF}: i=1,...,4\}$ in \cref{eq:lambda_134_MF,eq:lambda_2_MF}, followed by the practical fitting procedure used to obtain the corresponding coefficients according to the optimization in \cref{eq:theta_optim}.
This also includes selecting good initial values for $\ve{\theta}$ in order to facilitate convergence of the optimization.

The overall idea is to first train a GLaM using only the LF data to define both the starting points for the MF-GLaM coefficients and the basis functions that correspond to the LF expansion. 
Once these have been identified, the full HF and LF datasets are used to jointly maximize \cref{eq:llh}, thereby estimating the final coefficients $\ve c$ and $\ve d$.
We apply this procedure for multiple different configurations of the discrepancy truncation sets, with the best one identified via a modified Bayesian information criterion, described subsequently. 

More precisely, starting from the LF GLaM, we obtain both the truncation sets $\{\ca_i^\text{L}: i=1,...,4\}$ and initial coefficients $\{\ve c_{i, 0}: i=1,...,4\}$. 
We now address the truncation sets $\{\ca_i^\delta: i=1,...,4\}$ and initial coefficients $\{\ve d_{i, 0}: i=1,...,4\}$  associated with the discrepancy PCEs, starting by considering  a simplification regarding the relation between HF and LF GLaM shapes.
As discussed in \cref{sec:GLaM_single}, the shape of the PDF of a GLaM is primarily determined by the two parameters $\lambda_3$ and $\lambda_4$. 
In practice, we often find that the shape of the LF response PDF closely matches the corresponding HF shape.
Therefore, we set the discrepancies $\delta_3^\text{PC}, \delta_4^\text{PC}$ to zero ($\delta_3^{\mathrm{PC}}=\delta_4^{\mathrm{PC}}=0$) in \cref{eq:lambda_134_MF}, which gives us:
\begin{equation}
    \lambda_i^\text{MF} (\ve x) \approx 
    \sum_{\ua \in \ca_i^\text L} c_{i,\ua} \psi_{\ua}(\ve x), \quad i = 3,4.
\end{equation}
Let us note that this simplification is not a limitation of the framework, but rather a decision based on prior knowledge to reduce the overall MF model complexity, and therefore improve its data efficiency.
If prior knowledge or preliminary analysis suggests a significant discrepancy between the HF and LF distribution shapes, a low-order (e.g., degree 0 or 1) polynomial correction for  $\delta_3^\text{PC}$ and  $\delta_4^\text{PC}$ can be included instead.
Nonetheless, our empirical investigations suggest that while this adds flexibility, it often leads to slower convergence and reduced accuracy, particularly with limited HF data.

The expressions for $\lambda_1$ and $\lambda_2$ remain as in \cref{eq:lambda_134_MF} and \cref{eq:lambda_2_MF}, respectively.
To select the truncation sets $\{\ca_i^\delta: i=1,2\}$, we examine several candidate degrees 
$\ve p_{1, \delta} $ 
and 
$\ve p_{2, \delta}$,
and q-norms 
$\ve q_{1, \delta} $ 
and 
$\ve q_{2, \delta}$.
For each unique combination 
$\{(p_{1,\delta}^{(j)},q_{1,\delta}^{(k)},p_{2,\delta}^{(l)},q_{2,\delta}^{(m)})\}$, 
we initialize $\ve{d}_{i,0}\!=0$ (for $i=1,2$) and fit an MF-GLaM on all HF and LF data. 
Among the resulting models, we select the one that minimizes the following BIC score:
\begin{equation}\label{eq:BIC_MF}
    \text{MF-BIC} = -2 \, \ell(\ve{\theta};\cs, \ve{\cx},\cy_H, \cy_L) \, + \, n_{\ve \theta} \log \left ( \frac{N_\text L+N_\text H}{2} \right ).
\end{equation}
where $\ell(\ve{\theta};\cs, \ve{\cx},\cy_H, \cy_L)$ is the log-likelihood in \cref{eq:llh} and $n_{\ve \theta}$ is the total number of parameters in the set $\ve \theta$. 
A full Bayesian derivation of the MF-BIC score in \cref{eq:BIC_MF} is given in \cref{annex_B}.


The optimization strategy we use to obtain the parameters $\ve{\theta}$ that maximize the log-likelihood in \cref{eq:llh} is the same as the one described by \citet{ZhuIJUQ2020,ZhuSIAM2021}. More precisely,
the derivative-based trust-region method \citep{Steihaug1983} is applied first to solve the unconstrained problem. If constraints are violated (due to the dependence of the GLD support on the distribution parameters) we switch to the constrained (1+1)-CMA-ES algorithm \citep{Arnold2012} to obtain the final solution.

\section{Applications}
\label{sec:applications}
In this section, we investigate the performance of our multifidelity stochastic emulator across three examples of increasing complexity. 
The first example is an analytical toy example where both the HF and LF models are artificially defined GLaMs with four PCE-based parameters, whose coefficients are assigned manually.
The second example is a stochastic borehole model, where the HF model has three explicit input variables and five latent variables, whereas the LF model has only two explicit variables with six latent ones.
For our final example, we employ earthquake simulations of different fidelity to illustrate the approach in a more applied real-world setting.

We assess the accuracy of our stochastic emulators in terms of both their \emph{pointwise} approximation performance against the true HF simulator outputs, and their \emph{global} performance for approximating entire conditional distributions. We also examine the convergence behavior of our simulator for increasing high-fidelity training data. 

To assess the pointwise approximation capabilities of an MF-GLaM qualitatively, we plot the emulator response distributions at selected inputs throughout the input domain and compare them with the true/reference HF stochastic emulator response at these points.

To quantitatively assess the global performance of our emulators, traditional error measures used for deterministic emulators, such as the root mean squared error, are no longer appropriate, as the predictions of our stochastic emulators are full distributions rather than point values.
Instead, we use the mean-squared normalized Wasserstein distance of order two between the true HF stochastic model $Y_{\ve x}^\text H$, and our MF-GLaM emulator $\tilde Y_{\ve{x}}^\text{MF}$, defined as:
\begin{equation}
\label{eq:av_norm_Wass}
    \epsilon_\text{W} =\frac{\mathbb{E}_{\boldsymbol{X}}\left[d^2_{\mathrm{WS}}(Y_{\ve x}^\text H, \tilde Y_{\ve x}^\text{MF})\right]}{\text{Var}[Y^\text H]},
\end{equation}
where $\text{Var}[Y^\text H]$ is the total variance of the response of the true HF stochastic model, which accounts for both uncertainties in the input and the latent stochasticity, and $d_{\rm WS}$ is the \emph{Wasserstein distance of order two} \citep{Villani2009} between two random variables $Y_1, Y_2$ with inverse CDF $Q_1, Q_2$:
\begin{equation}
\label{eq:Wass_dist}
    d_{\mathrm{WS}}\left(Y_1, Y_2\right)=\left\|Q_1-Q_2\right\|_2=\sqrt{\int_0^1\left(Q_1(u)-Q_2(u)\right)^2 \mathrm{~d} u}.
\end{equation}

The expectation in \cref{eq:av_norm_Wass} is approximated using the sample average over a test set $\ve{\cx}_\text{test} = \{\ve{x}_\text{test}^{(1)}, ..., \ve{x}_\text{test}^{(N_\text{test})}\}$ of size $N_\text{test} = 1,\!000$, generated via Latin hypercube sampling (LHS) \citep{McKay1979}.
In the first application, where the true HF model is itself a GLaM, the quantiles to compute the Wasserstein distance in \cref{eq:Wass_dist} are obtained directly from \cref{eq:GLD quantile}.
Furthermore, the variance in the denominator of \cref{eq:av_norm_Wass} is estimated using the law of total variance, expressed as:
\begin{equation}
    \text{Var}[Y_\text H]=\mathbb{E}_{\boldsymbol{X}}\left[ \text{Var}(Y^\text{H}_{\ve{x}})\right]
        + \text{Var}\left(\mathbb{E}_{\boldsymbol{X}}[Y^\text{H}_{\ve{x}}]\right).
\end{equation}
This variance is estimated on $\ve{\cx}_\text{test}$, where the conditional means $\mathbb{E}_{\boldsymbol{X}}[Y^\text{H}_{\ve{x}}]$ and variances $\text{Var}(Y^\text{H}_{\ve{x}})$ at each point are obtained from \cref{eq:GLD_mean,eq:GLD_var}.
In contrast, for the final two applications, where neither the analytical form of the true HF response distribution $Y_{\ve x}^\text H$ nor its moments and quantiles are available, we perform $N_\text{R}=10^4$ replications (i.e., repeated evaluations) of the HF stochastic simulator at each $\ve{x}_\text{test}^{(i)}$. 
Then, the quantiles of each response distribution are approximated by empirical quantiles across the replications, while the overall variance $\text{Var}[Y_\text H]$ is estimated as the empirical variance of the aggregated responses across all $N_\text{R} \times N_\text{test}$ evaluations.

Moreover, as part of our validation procedure, we aim to assess the added value of the MF-GLaM in comparison to single-fidelity models, both in terms of local and global performance.
We also investigate whether the multifidelity emulator achieves a faster convergence rate in terms of number of HL model evaluations.
For comparison purposes, alongside each MF-GLaM, we construct two corresponding single-fidelity GLaMs: one trained solely on the HF data used in the MF-GLaM, referred to as the \quot{HF-only} GLaM, and another trained exclusively on the associated LF data, referred to as the \quot{LF-only} GLaM.

The global performance of the stochastic emulator with respect to the HF experimental design size is investigated by performing simulations with increasing HF sample sizes, $N_\text{H} = \{100, 200, 400, 800\}$, while keeping the size of the LF experimental design fixed to $N_\text{L} = 1,\!000$ data points.
To account for statistical variability in the experimental designs,  $N_\text{r} = 25$ independent HF and LF designs are generated for each configuration, and the corresponding GLaMs are constructed.
An aggregated view of the results is provided using box plots.

To qualitatively assess the pointwise approximation capabilities of all stochastic emulators, we select, for each HF experimental design size, the HF, LF, and MF GLaMs corresponding to the \emph{median} Wasserstein distance $\epsilon_\text{W}$ among the $N_\text{r} = 25$ repetitions. 
Their predicted response distributions are then visualized at selected input locations and compared against the true (reference) HF stochastic simulator responses.

Across all three examples, the candidate PCE degrees and q-norms for the polynomial expansion of each parameter in the HF-only, LF-only and MF- GLaMs, are specified as shown in \cref{tab:expansions_summary}.
When $N_{\mathrm{H}}=100$, the HF and discrepancy degrees are restricted to mitigate the risk of overfitting. 
Finally, the implementation of our code for the MF-GLaM methodology and its associated applications is built upon the existing GLaM module of UQLab version 2.1.0 \citep{Marelli2014,UQdoc_21_120_GLaM}, a general-purpose uncertainty quantification software developed in Matlab.

\begin{table}[htb]
\centering
\caption{Summary of candidate PCE degrees and q-norms for the parameters in the HF and LF GLaMs, and for the discrepancy expansions in the MF-GLaM. 
}
\label{tab:expansions_summary}
\begin{tabular}{lcccccc}
\toprule
\multirow{2}{*}{{Param.}}  
& \multicolumn{2}{c}{{HF\textsuperscript{*}/LF Expansions}} & \multicolumn{2}{c}{{Discrepancy Expansions\textsuperscript{\dag}}} \\
\cmidrule(lr){2-3} \cmidrule(lr){4-5}
& {Degrees} & {q-norms} 
& {Degrees} & {q-norms}\\
\midrule
$\lambda_1^{\mathrm{PC}}$
& $\{1,\dots,6\} $ 
& $\{0.2,0.4,0.6,0.8,1\} $
& $\{0,\dots,2\}$ 
& $\{0.6,1\}$ \\
$\lambda_2^{\mathrm{PC}}$
& $\{1,\dots,4\} $
& $\{0.2,0.4,0.6,0.8,1\} $
& $\{0,1\}$
& $\{1\}$ \\
$\lambda_3^{\mathrm{PC}}$
& $\{0,\dots,2\} $
& $\{0.6,1\} $
& -- & -- \\
$\lambda_4^{\mathrm{PC}}$
& $\{0,\dots,2\} $
& $\{0.6,1\} $
& -- & -- \\
\bottomrule
\end{tabular}
\vspace{1em}
\begin{minipage}{0.9\textwidth}
\small
\textsuperscript{*}When $N_{\mathrm{H}}=100$, the HF expansion degrees and q-norms are reduced: \newline
\(
    \ve{p}_{1,\text{H}}=\{1,2\}, \:
    \ve{p}_{2,\text{H}}=\{1\}, \:
    \ve{p}_{3,\text{H}}=\ve{p}_{4,\text{H}}=\{0,1\}
    \:\text{and}\: 
    \ve{q}_{1,\text{H}}=\{0.6, 1\}, \:
    \ve{q}_{2,\text{H}}=\ve{q}_{3,\text{H}}=\ve{q}_{4,\text{H}}=\{1\}
    .
\)
\newline
\textsuperscript{\dag}When $N_{\text{H}}=100$, the discrepancy expansion degrees and q-norms are reduced: \newline
\(
\ve{p}_{1,\delta}=\ve{p}_{2,\delta}=\{0,1\}
\quad\text{and} \quad \ve{q}_{1,\delta}=\ve{q}_{2,\delta}=\{1\}.
\)
\end{minipage}
\end{table}

\subsection{Synthetic GLaM models}
In this first example, both the high‑fidelity and low‑fidelity stochastic simulators are GLaMs defined on a four‑dimensional input space.
As the true response distributions exactly follow the GLaM form, this case provides the minimal‑complexity setting needed to validate our multifidelity procedure in the case of perfect model response agreement, before proceeding to more complex, imperfectly represented models.
The input variables $X_1, X_2, X_3, X_4$ are assumed to follow independent uniform distributions over $[0,2]$, i.e. $X_1, X_2, X_3, X_4 \sim \mathcal{U}([0, 2])$.  
The four parameters $\ve{\lambda}^\text{H}$ and $\ve{\lambda}^\text{L}$ of the HF and LF model, respectively, are represented using polynomial chaos expansions, with the corresponding nonzero coefficients listed in \cref{tab:synth_GLaM_coeffs}.
Each multi-index defines a polynomial basis function, where a value $p$ at the $i$-th position of a given multi-index corresponds to $x_i$ raised to the power of $p$. 
For example, the multi-index 2100 corresponds to the product of two Legendre polynomials: one of order $2$ in $x_1$, and one of order $1$ in $x_2$.
\begin{table}[htb]
\centering
\caption{Non-zero coefficients of the expansions of the parameters $\ve{\lambda}$ of the true HF and LF GLaMs}
\label{tab:synth_GLaM_coeffs}
\begin{tabular}{c cc cc cc cc}
\toprule
Multi-index 
& \multicolumn{2}{c}{$\lambda_1$}
 & \multicolumn{2}{c}{$\lambda_2$}
 & \multicolumn{2}{c}{$\lambda_3$}
 & \multicolumn{2}{c}{$\lambda_4$} \\
\cline{2-9}
 & HF & LF
 & HF & LF
 & HF & LF
 & HF & LF \\
\midrule
0000 & 2      & 2.2    & 1.2    & 0.5   & 0.38 & 0.35 & 0.4  & 0.42 \\
0001 & 3.5    & 2      & $-1.1$ & $-0.1$ & --    & --   & --   & --   \\
0010 & 2.45   & 3      & --     & --     & 0.2   & 0.2  & --   & --   \\
0100 & $-0.5$ & $-0.3$ & 0.3    & 1    & --    & --   & --   & --   \\
1000 & 0.2    & --     & 0.8    & --     & --    & --   & --   & --   \\
0020 & 0.05   & --     & --     & --     & --    & --   & --   & --   \\
0200 & 2.3    & 2      & --     & --     & --    & --   & --   & --   \\
2000 & 2.3    & 2.5    & --     & --     & --    & --   & --   & --   \\
0011 & 0.12   & --     & --     & --     & --    & --   & --   & --   \\
1100 & 0.5    & --     & --     & --     & --    & --   & --   & --   \\
2100 & 0.04   & 0.041  & --    & --     & --    & --   & --   & --   \\
1110 & 0.02   & 0.022  & --    & --     & --    & --   & --   & --   \\
\bottomrule
\end{tabular}
\end{table}

\begin{figure}[htb]
    \centering
    \includegraphics[width=\textwidth,clip=true,trim=10 0 80 0]{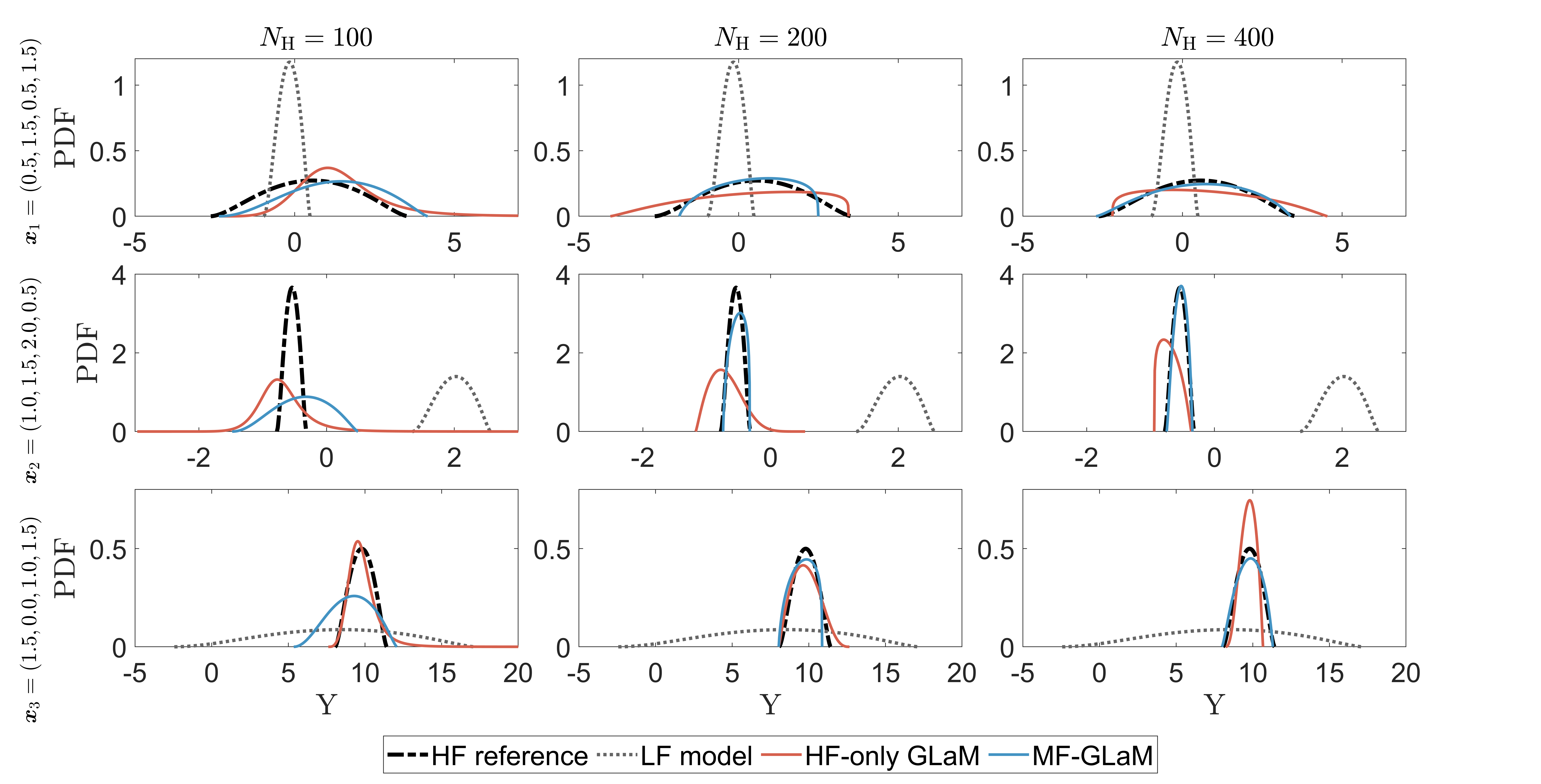}
    \caption{Synthetic GLaMs example. Comparison of the true HF (black dashed curve) and LF (black dotted curve) conditional PDFs and the corresponding predictions from an MF-GLaM (blue curve) and a GLaM trained on the HF data only (red curve). The comparison is made for three representative points and for increasing size of HF training data.}
    \label{fig:06Synth_PointApprox}
\end{figure}

To illustrate how the HF and LF models differ across the input domain, \cref{fig:06Synth_PointApprox} compares their conditional response distributions at three representative input points. In this figure, HF responses are shown as black dashed curves, while LF responses appear as gray dotted curves.
The same plots also compare the predictions from two stochastic emulators: an MF-GLaM (blue curves) and a GLaM trained solely on HF data (red curves), the latter referred to as the “HF-only” GLaM. Results are shown for increasing HF training set sizes, ranging from $N_\text{H} = 100$ up to $N_\text{H} = 400$ data points, while the LF training set size remains fixed at $N_\text{L} = 1,\!000$ data points.
We notice that, for all three input points and across all HF training set sizes, the MF-GLaM consistently captures the conditional HF distribution more accurately than the HF-only GLaM. 
Even for $\ve{x}_2$, where the LF and HF response supports are disjoint, a quite uncommon scenario in practice, the MF‑GLaM reproduces the HF conditional PDF more faithfully compared to relying solely on HF data.
We notice that in this four-dimensional application, $N_\text{H} = 100$ HF training data points are insufficient to achieve an acceptable approximation of the response PDF; however, the MF approach still yields improved predictions compared to the HF-only model. Once the HF training set size increases to $N_\text{H} = 200$, the MF predictions become very close to the true HF PDF, and with $N_\text{H} = 400$, the MF approximation is nearly exact.
In contrast, the HF-only GLaM exhibits greater variability and struggles to reproduce the PDF shape in certain cases (particularly for inputs $\ve{x}_1$ and $\ve{x}_2$).

\begin{figure}[htb]
    \centering
    \includegraphics[width=0.6\textwidth]{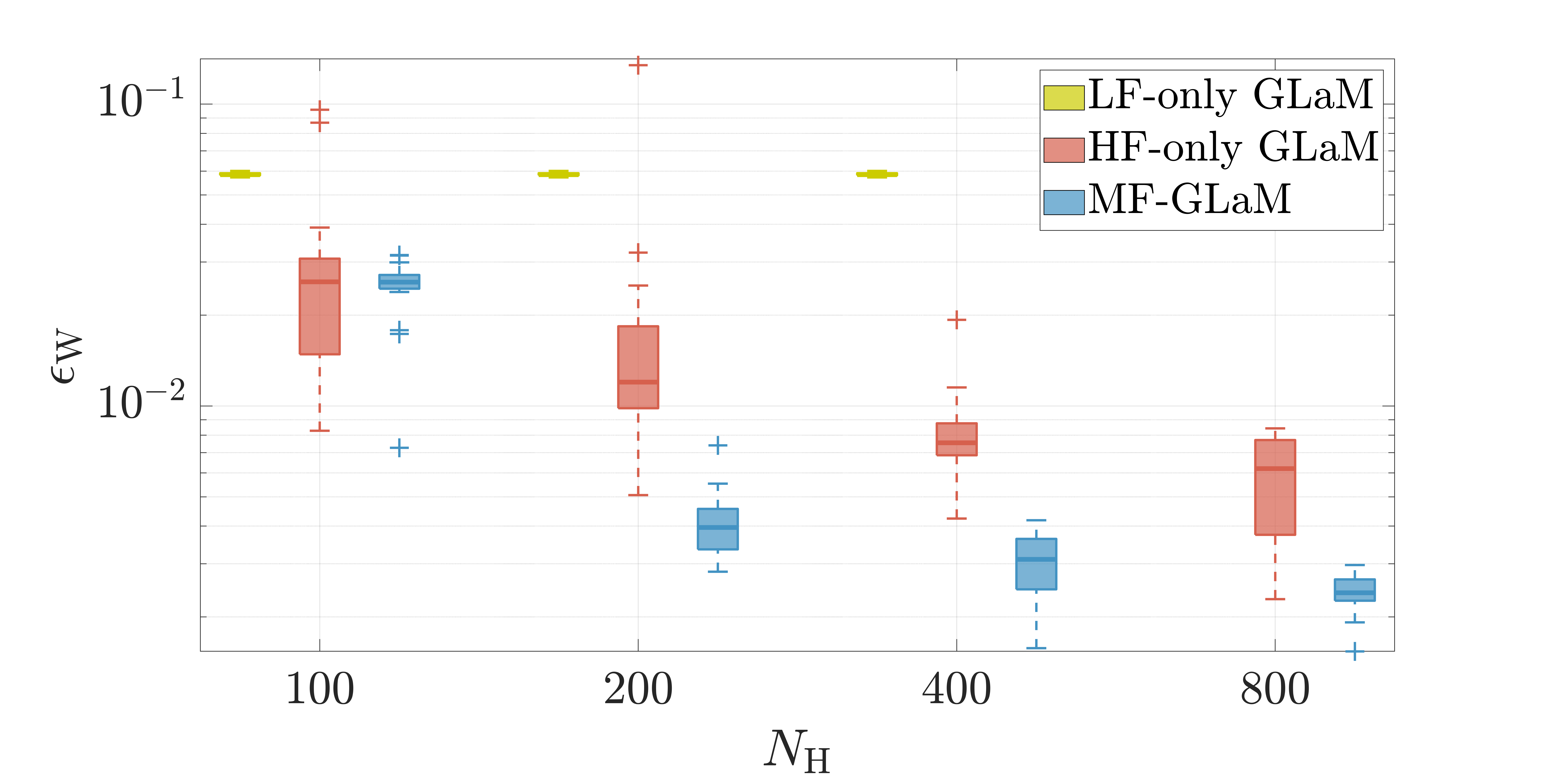}
    \caption{Synthetic GLaMs example. Comparison of the normalized Wasserstein distance $\epsilon_\text{W}$ for LF-only, HF-only, and MF GLaMs across increasing HF experimental design sizes $N_\text{H}$.
    The box plots correspond to $N_\text{r}=25$ repetitions of the procedure, and $\epsilon_\text{W}$ is computed using $1,\!000$ test points.
    }
    \label{fig:06Synth_WassDist}
\end{figure}

These observations are further supported by the global approximation performance comparison shown in \cref{fig:06Synth_WassDist}, and its numerical summary, provided in \cref{tab:06Synth_WassDist_summary}.
The box plot presents the performance comparison in terms of normalized Wasserstein distance $\epsilon_\text{W}$ of the MF-GLaM, the HF-only GLaM, and the LF-only GLaM, the latter serving as a baseline.  
Each box group corresponds to a different HF training set size, ranging from $N_\text{H} = 100$ to $N_\text{H} = 800$ points, with $N_\text{r}=25$ repetitions per scenario. Throughout these tests, we drew once $N_\text{r}$ LF training datasets of size $N_\text{L} = 1,\!000$ each, which were reused across all $N_\text{H}$ values.

We observe that the MF-GLaM consistently outperforms the GLaM trained solely on the HF data and shows faster convergence to the true HF model, indicated by the rapid decrease in Wasserstein distance. 
Especially for $N_\text{H} = 200$, the error is reduced by approximately one order of magnitude.
The only exception occurs at $N_\text{H} = 100$, where the median performance of the MF and HF-only emulators appears comparable.
This can be attributed to the imposed limitation of the correction term $\lambda_1^{\delta, \text{PC}}$ to a polynomial degree of 1 at this training size, introduced to avoid overfitting. 
However, as shown in \cref{tab:synth_GLaM_coeffs}, the actual discrepancy between the HF and LF expansions of $\lambda_1$ involves higher-degree terms.
Nonetheless, the MF-GLaM demonstrates reduced variability across repetitions, as indicated by the significantly smaller interquartile range of $\epsilon_\text{W}$ in \cref{tab:06Synth_WassDist_summary}, which suggests more robust predictive performance.

\begin{table}[htb]
\centering
\caption{
Synthetic GLaMs example. Median (interquartile range) of the normalized Wasserstein distance $\epsilon_\text{W}$ $\times10^{-2}$, computed over $N_\text{r}=25$ repetitions, comparing HF‑only GLaM and MF‑GLaM models for four HF training set sizes.
Baseline $\epsilon_\text{W}$, for the LF‑only GLaM: $5.8 \, (0.1) \times 10^{-2}$.}
\label{tab:06Synth_WassDist_summary}
\begin{tabular}{c c c}
\toprule
$N_H$ & HF-only GLaM & MF-GLaM \\
\midrule
100  & 2.6\,(1.6) & 2.6\,(0.3) \\
200  & 1.2\,(0.9) & 0.4\,(0.1) \\
400  & 0.8\,(0.2) & 0.3\,(0.1) \\
800  & 0.6\,(0.4) & 0.2\,(0.04) \\
\bottomrule
\end{tabular}
\end{table}

Finally, we also compare how well the emulators reproduce the \emph{mean} and \emph{variance} of the HF stochastic model. Since these two statistical quantities are deterministic, we use a normalized mean‐square error (NMSE) measure to assess their prediction accuracy. Specifically, for each point $\ve{x}^{(i)}$ in a test set of size $N_{\mathrm{t}} = 10^4$, the NMSE is given by:
\begin{equation}\label{eq:val_error_NRMSE}
    \epsilon_\text{v} =\frac{\sum_{i=1}^{N_\text{t}} (\eta_\text{GLaM}^{(i)}-\eta_\text{H}^{(i)})^2}{\sum_{i=1}^{N_\text{t}} (\eta_\text{H}^{(i)}-\hat{\mu}_\eta)^2} , 
    \quad \text{where} \quad
    \hat{\mu}_\eta = \frac{1}{N_\text{t}} \sum_{i=1}^{N_\text{t}} \eta_\text{H}^{(i)},
\end{equation}
and $\eta_\text{H}^{(i)}$ the true value of the mean or variance at $\ve{x}^{(i)}$ for the HF simulator, given by \cref{eq:GLD_mean} and \cref{eq:GLD_var}, respectively, since in this example our true HF simulator is a GLaM.
Also, $\eta_\text{GLaM}^{(i)}$ is the corresponding mean or variance value predicted by the GLaM surrogate model.

\begin{figure}[htb]
     \centering
     \subfigure[]{
         \includegraphics[width=.475\textwidth]{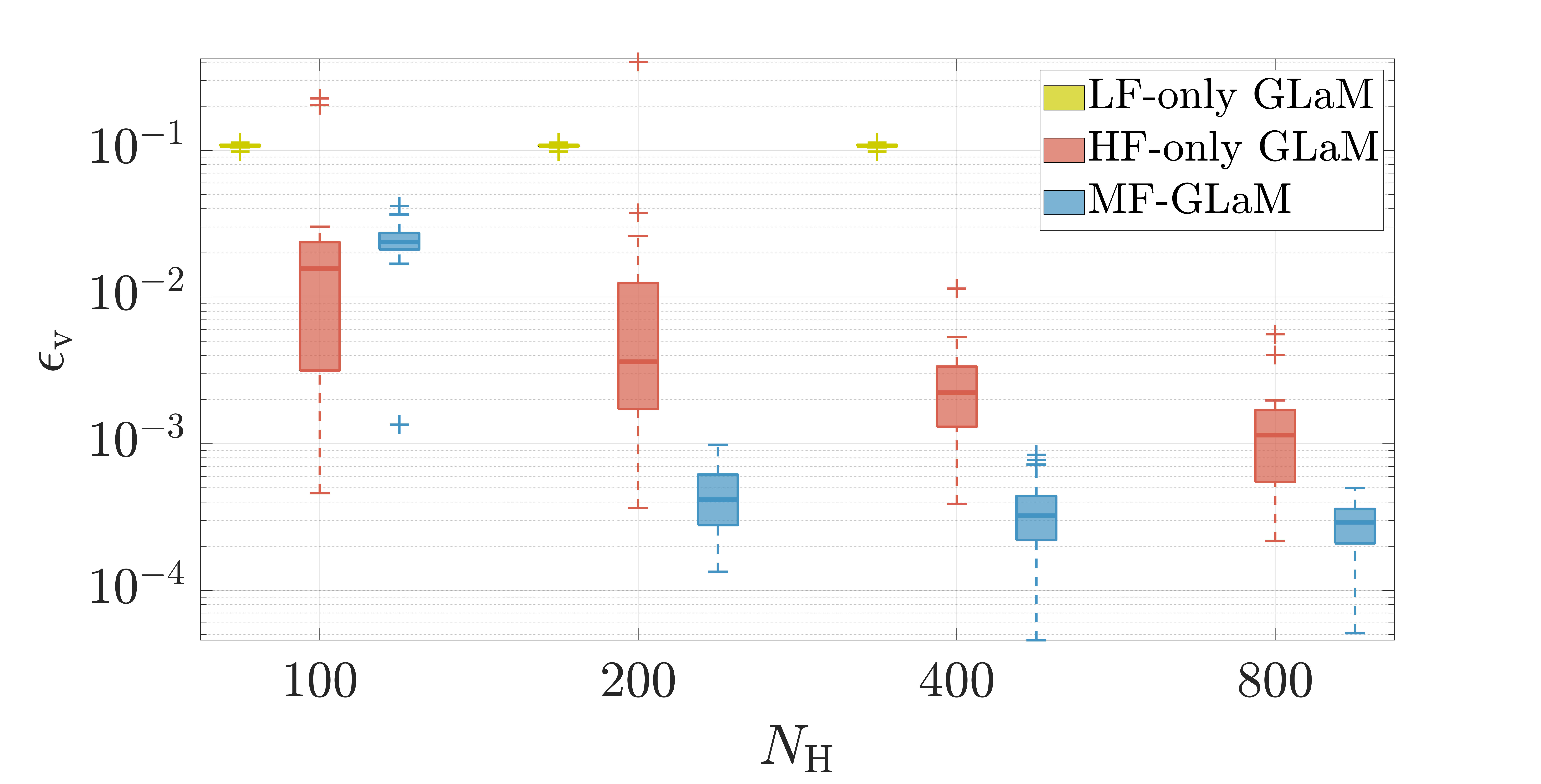}
         \label{fig:06Synth_RMSE_mean}
     }%
    \hfill
    \subfigure[]{
         \includegraphics[width=.475\textwidth]{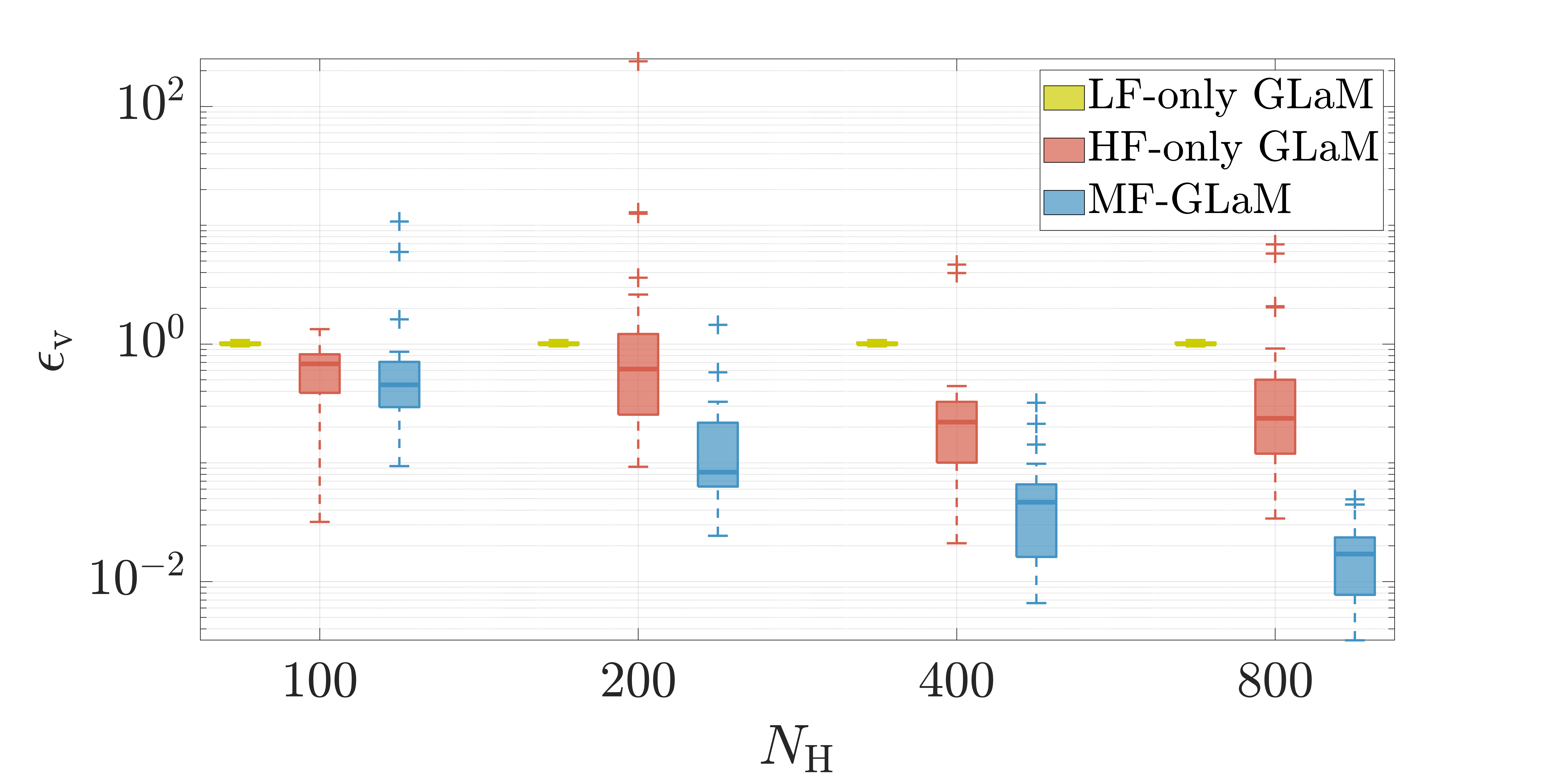}
         \label{fig:06Synth_RMSE_var}
    }

    \caption{Synthetic GLaMs example. Comparison of the normalized mean-squared error $\epsilon_\text{v}$ on the prediction of (a) the mean and (b) the variance, for LF-only, HF-only, and MF GLaMs across increasing HF experimental design sizes $N_\text{H}$.
    The box plots correspond to $N_\text{r}=25$ repetitions of the procedure, and $\epsilon_\text{v}$ is computed using $N_\text{t} = 10^4$ test points.}
    \label{fig:06Synth_RMSE_mean_var}
\end{figure}

\cref{fig:06Synth_RMSE_mean_var} shows the NMSE for the mean (\cref{fig:06Synth_RMSE_mean}) and variance (\cref{fig:06Synth_RMSE_var}) estimates of the stochastic emulators whose predictive distributions were evaluated in \cref{fig:06Synth_WassDist}. 
The corresponding box plots align with those for the Wasserstein distance and further confirm that the MF-GLaM provides more accurate and robust predictions compared to its single-fidelity counterparts, also for the statistics of the model response.

\subsection{Stochastic borehole function}
Our second application example is based on a well-known engineering benchmark function, the borehole function, which models the water flow through a borehole drilled from the ground surface through two aquifers. 
The water flow rate, in \(m^3/yr\) is determined by the equation:
\begin{equation}
\label{eq:borehole_det}
    f(\ve{x}) = \frac{2\pi \, t_u (h_u-h_l)}
    {\ln{(r/r_w)}(1 + \frac{2lt_u}{\ln{(r/r_w)}r_w^2 k_w} + \frac{t_u}{t_l})}
\end{equation}
The input vector $\ve{x}$ contains eight random variables, whose distributions are shown in \cref{tab:borehole_inputvars}.
\begin{table}[htb]
\centering
\caption{Stochastic borehole example. Input variables and their distributions}
\label{tab:borehole_inputvars}
    \begin{tabular}{lll}
        \toprule
        Variable & Distribution & Description \\ 
        \midrule
        $R_w$ [m]& $\mathcal{N} \,([0.1,\; 0.016])$ &radius of borehole\\
        $H_u$ [m]& $\mathcal{U} \,([990,\; 1110])$ &potentiometric head of upper aquifer\\
        $K_w$ [m/yr]& $\mathcal{U} \,([9855,\; 12045])$  &hydraulic conductivity of borehole\\
        $R$   [m]& $\mathcal{LN} \,([7.71,\; 1.0056])$ &radius of influence\\
        $T_u$ [$\text{m}^2$/yr]& $\mathcal{U} \,([63070,\; 115600])$ &transmissivity of upper aquifer\\
        $T_l$ [$\text{m}^2$/yr]& $\mathcal{U} \,([63.1,\; 116])$ &transmissivity of lower aquifer\\
        $H_l$ [m]& $\mathcal{U} \,([700,\; 820])$  &potentiometric head of lower aquifer\\
        $L$   [m]& $\mathcal{U} \,([1120,\; 1680])$ &length of borehole\\
        \bottomrule
    \end{tabular}
\end{table}

The borehole function was first used in a deterministic multifidelity context by \citet{Xiong2013}, who consider as the LF model the following modified function:
\begin{equation}
\label{eq:borehole_det_LF}
    f_\text{L}(\ve{x}) = \frac{5 \, t_u (h_u-h_l)}
    {\ln{(r/r_w)}(1.5 + \frac{2lt_u}{\ln{(r/r_w)}r_w^2 k_w} + \frac{t_u}{t_l})} .
\end{equation}

Moreover, \citet{LuethenCMAME2023} modified the borehole function in \cref{eq:borehole_det} into a stochastic simulator by treating the five input parameters, $\ve{\Omega}_{\text{H}}=(R,T_u,T_l,H_l,L)$, as latent variables.
We adopt this variation for our application as the HF stochastic simulator, which results in the following three-dimensional HF model: 
$f_{s, \text{H}}(r_w, h_u, k_w) = f(r_w, h_u, k_w; \omega_\text{H})$. 

Additionally, we obtain the LF stochastic simulator from the deterministic LF model in \cref{eq:borehole_det_LF} by considering six of its input parameters to be latent: $\ve{\Omega}_\text{L} = (K_w, R, T_u, T_l, H_l, L)$. This results in the following two-dimensional LF model:
$f_{s, \text{L}}(r_w, h_u) = f_\text{L}(r_w, h_u; \omega_\text{L})$. 

This example investigates the applicability of the multifidelity GLaM methodology in cases where the LF explicit input variables are a subset of the HF ones. 
In this example, the two-dimensional LF input is a subset of the three-dimensional HF input.
As a result, whereas the two common input variables can be modeled using all the available HF and LF data in the multifidelity GLaM through both the LF and the discrepancy part in \cref{eq:lambda_MF}, the additional HF-only input variable and its interactions have to be captured exclusively through the discrepancy component of the MF-GLaM, relying solely on the HF data.

\begin{figure}[htb]
    \centering
    \includegraphics[width=\textwidth,clip=true,trim=10 0 80 0]{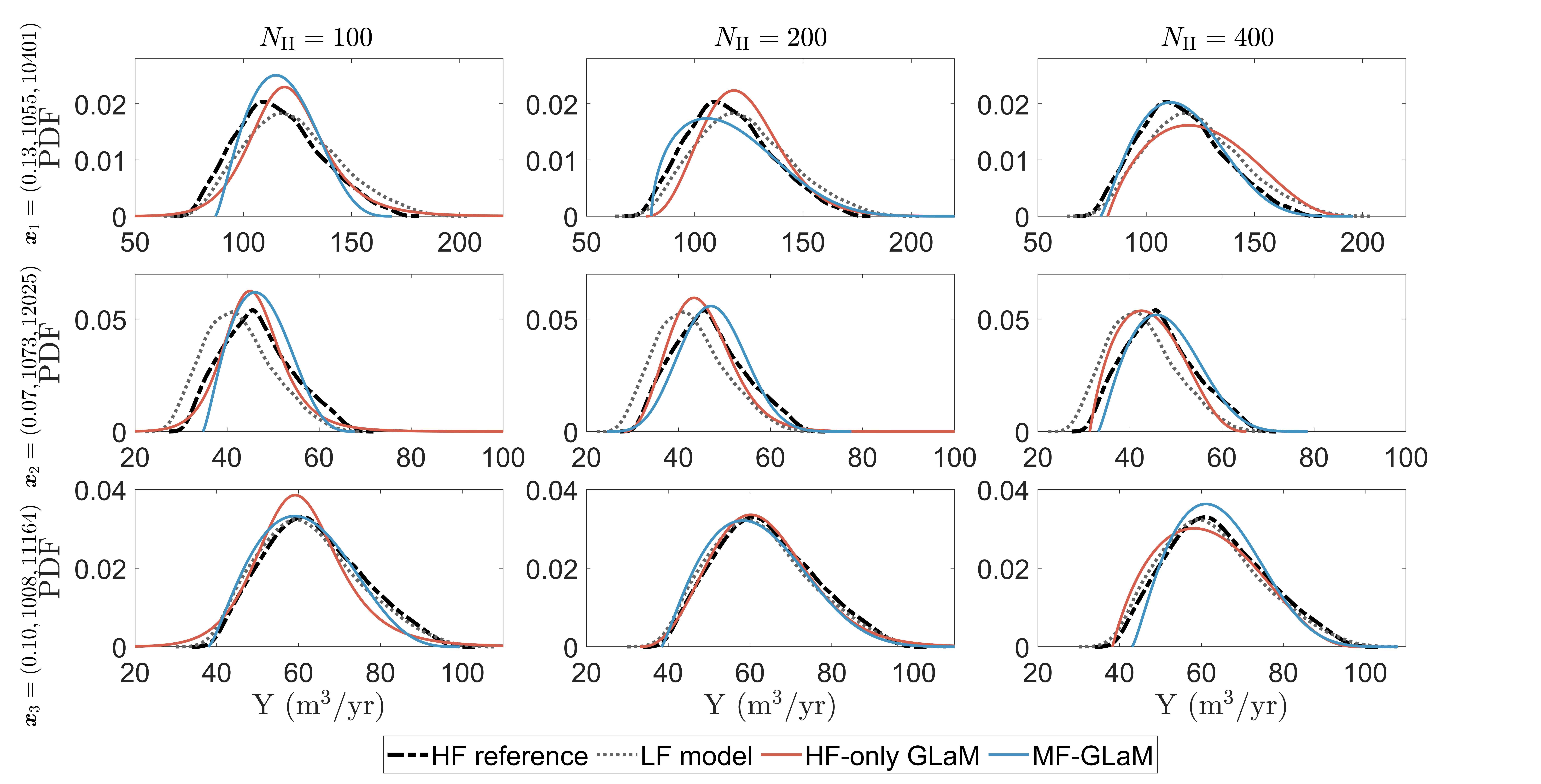}
    \caption{Stochastic borehole example. Comparison of the true HF (black dashed curve) and LF (black dotted curve) conditional PDFs and the corresponding predictions from an MF-GLaM (blue curve) and a GLaM trained on the HF data only (red curve). The comparison is made for three random points and for increasing size of HF training data.}
    \label{fig:02Borehole_PointApprox}
\end{figure}

\cref{fig:02Borehole_PointApprox} compares the PDF approximation capabilities of an MF-GLaM and an HF-only GLaM for three random points. In the same plots, we see the true HF PDF (black dashed curve) as well as the true LF (gray dotted curve) PDF. We notice that the HF-LF relationship can vary across the input space. For example, at $\ve{x}_3$, the LF PDF closely matches the HF reference, whereas at $\ve{x}_1$ and $\ve{x}_2$ the LF distribution is more noticeably shifted or differs in shape. The MF-GLaM (blue curve) successfully corrects for these discrepancies, consistently aligning with the true HF response, even when $N_\text{H} = 100$.
In contrast, the HF-only GLaM (red curve) tends to misplace the peak of the PDF or produce much longer tails, especially at smaller training sizes. 

\begin{figure}[htb]
    \centering
    \includegraphics[width=0.6\textwidth]{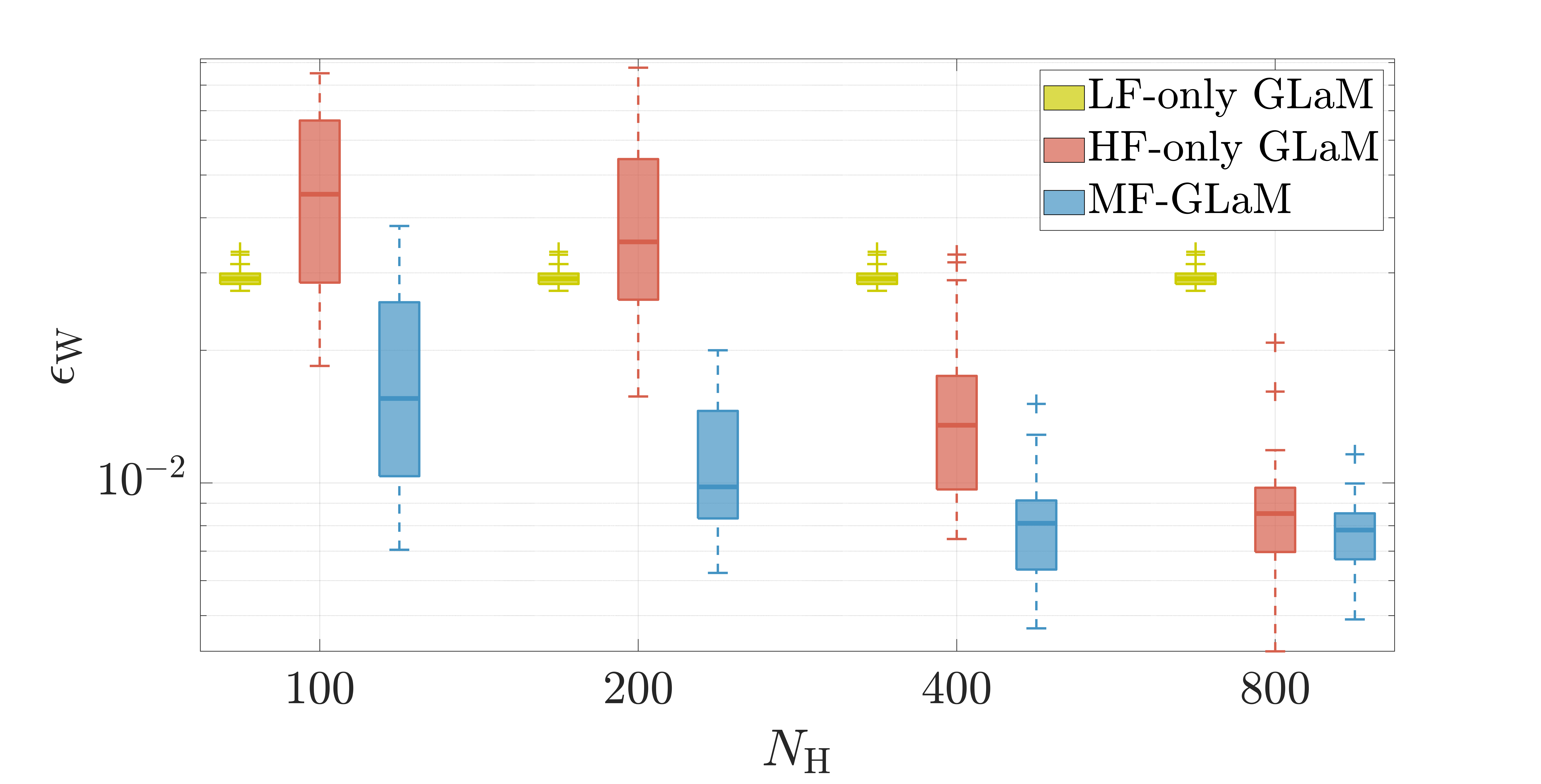}
    \caption{Stochastic borehole example. Comparison of the normalized Wasserstein distance $\epsilon_\text{W}$ for LF-only, HF-only, and MF GLaMs across increasing HF experimental design sizes $N_\text{H}$.
    The box plots correspond to 25 repetitions of the procedure, and $\epsilon_\text{W}$ is computed using $1,\!000$ test points.}
    \label{fig:02Borehole_WassDist}
\end{figure}

\cref{fig:02Borehole_WassDist} and its numerical summary in \cref{tab:02Borehole_WassDist_summary} compare the global approximation capabilities of the LF‐only, HF‐only, and MF‐GLaMs in terms of the normalized Wasserstein distance $\epsilon_\text{W}$.
The size of the HF training set ranges from $N_\text{H} = 100$ up to $N_\text{H} = 800$, while the LF training set size is fixed at $N_\text{L} = 1,\!000$ data points.
Overall, the MF‐GLaM (blue boxes) consistently outperforms the HF‐only GLaM (red boxes), especially when the HF training set is small. Particularly, with only $N_\text{H} = 100$ high‐fidelity samples, the MF-GLaM achieves accuracy comparable to that of the HF‐only model trained with $N_\text{H} = 400$. 
As $N_\text{H}$ increases to 800, the gap between the MF and HF‐only approaches narrows; however, such a large high‐fidelity to low-fidelity data size ratio is generally atypical for multi‐fidelity scenarios.

\begin{table}[htb]
\centering
\caption{
Stochastic borehole example. Median (interquartile range) of the normalized Wasserstein distance $\epsilon_\text{W}$ $\times10^{-2}$, computed over $N_\text{r}=25$ repetitions, comparing HF‑only GLaM and MF‑GLaM models for four HF training set sizes.
Baseline $\epsilon_\text{W}$, for the LF‑only GLaM: $2.9 \, (0.2) \times 10^{-2}$.
}
\label{tab:02Borehole_WassDist_summary}
\begin{tabular}{c c c}
\toprule
$N_H$ & HF-only GLaM & MF-GLaM \\
\midrule
100  & 4.5 (3.8) & 1.6 (1.5) \\
200  & 3.5 (2.8) & 1.0 (0.6) \\
400  & 1.4 (0.8) & 0.8 (0.3) \\
800  & 0.9 (0.3) & 0.8 (0.02) \\
\bottomrule
\end{tabular}
\end{table}

The normalized root-mean-squared errors for approximating the mean and variance of the response distributions confirm the same trends. 
Hence, we omit the corresponding box plots here and in the subsequent example, for brevity.

\subsection{Multi-story building subject to an earthquake}
In our last application, we aim to explore the applicability and performance of our framework in a real-world application involving real earthquake simulations, adopted from \citet{ZhuPEM2023,Schear2025}.
More precisely, we consider a three‐story steel frame structure subjected to stochastic seismic excitations.
The structure is modeled as a three-degree-of-freedom system and is represented by a computational model, and the input is given by synthetic ground motion time series generated from a stochastic ground motion model fitted to a dataset of real ground motions.
In this framework, the quantity of interest (QoI) is the maximum interstory drift, defined as the maximum of the individual interstory drifts, $\Delta_1(t), \Delta_2(t), \Delta_3(t)$, computed for each story during the seismic event: 
\begin{equation}
    Y = \max_{i \in\{1,2,3\}} \; \max_t \Delta_i (t).
\end{equation}

\begin{figure}[htbp]
  \centering
  \includegraphics[width=0.6\textwidth]{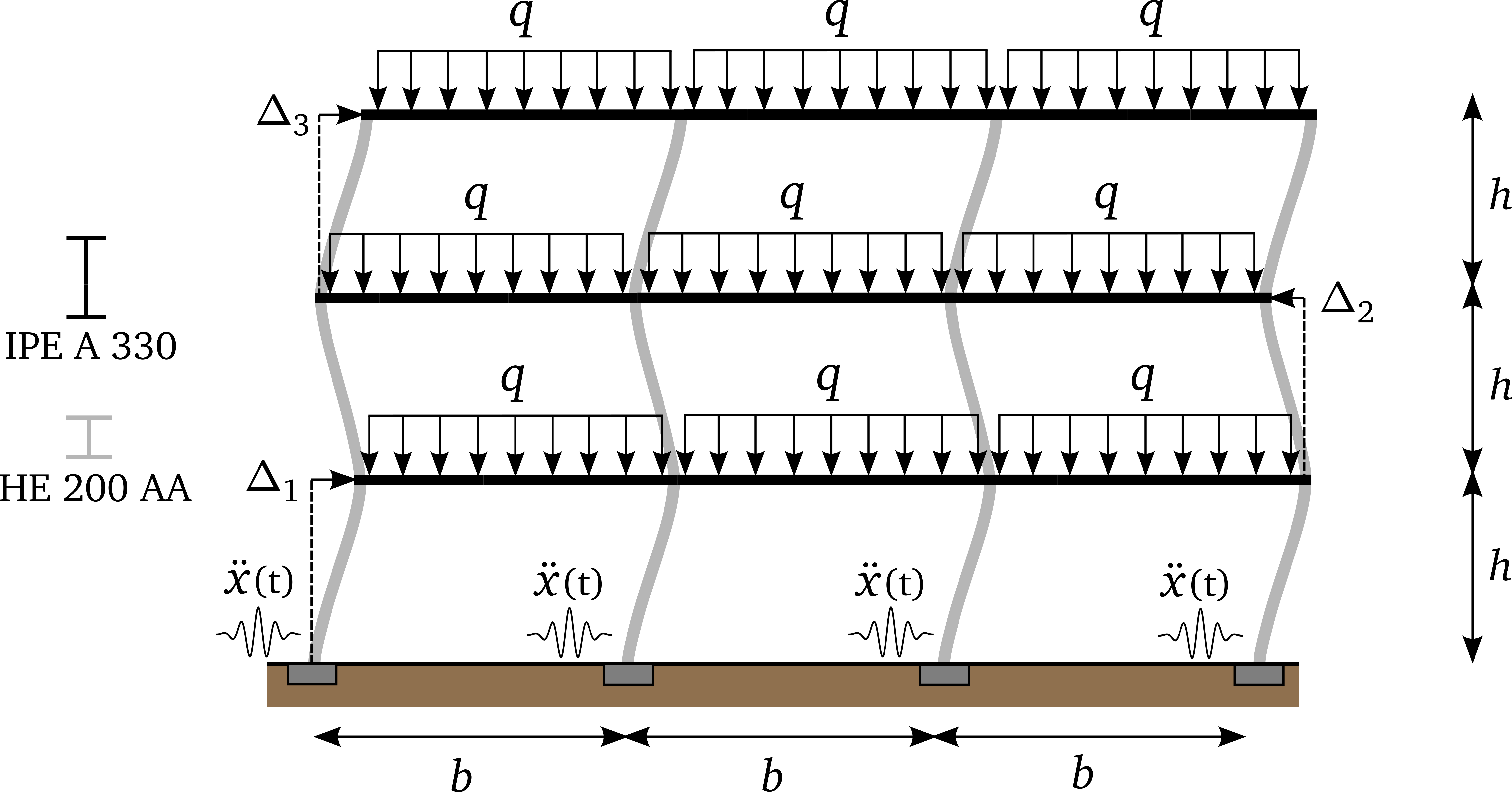}
  \caption{%
    Illustration of the three-story frame structure.
    Figure reproduced from \citet{Schear2025} \emph{with permission}.
  }
  \label{fig:gld_shape}
\end{figure}

For the synthetic ground motion model, we adopt the model from \citet{ZhuPEM2023}, based on the models from \citet{Broccardo2017,Rezaeian2010}. The uncertain seismic input is characterized by four ground motion parameters: the expected Arias intensity $I_a$, the time $t_{\text{mid}}$ at which 45\% of the expected Arias intensity is reached, the effective duration $D_{5-95}$ of the ground motion, and the main frequency $\omega_g$ of the motion at $t_{\text{mid}}$. 
The ground motion parameters are statistically modeled using log-normal marginal distributions coupled with a Gaussian copula, with the details of the distributions shown in \cref{tab:earthquake_inputvars}.
As the four ground‐motion parameters are high‐level descriptors of an earthquake signal, they do not uniquely define a full time‐series ground motion. Thus, a single parameter set can lead to significantly different possible signals.

\begin{table}[htb]
\centering
\caption{Multi-story building subject to an earthquake. Ground motion parameters and the correlation matrix of their Gaussian copula.}
\label{tab:earthquake_inputvars}
\begin{tabular}{l l c}
\toprule
{Name} & {Distribution} & {Correlation matrix $R$} \\
\midrule
$I_a$ [m/s]    
  & $\mathcal{LN}\bigl($-4.61$,\;2.0920\bigr)$
  & \multirow{4}{*}{$
  \begin{pmatrix}
    1 & 0.015 & $-0.23$ & $-0.13$\\
    0.015 & 1 & 0.68 & $-0.36$\\
    $-0.23$ & 0.68 & 1 & $-0.11$\\
    $-0.13$ & $-0.36$ & $-0.11$ & 1
  \end{pmatrix}
  $}\\
$t_{\mathrm{mid}}\ [\text{s}]$ 
  & $\mathcal{LN}\bigl(2.55,\;0.7998\bigr)$
  & \\
$D_{5\text{--}95}\ [\text{s}]$ 
  & $\mathcal{LN}\bigl(2.67,\;0.2826\bigr)$
  & \\
$\omega_g\ [\text{rad/s}]$ 
  & $\mathcal{LN}\bigl(1.42,\;0.3475\bigr)$
  & \\
\bottomrule
\end{tabular}
\end{table}

The three-story steel frame is modeled as a three-degree-of-freedom shear-type system with a Bouc--Wen hysteretic law governing the inelastic force-displacement relationship \citep{Bouc1967,Wen1976}.
Each story is assumed to have a height of $h=3$~m and a floor width of $b=5$~m, with beams subjected to a uniformly distributed load of $q=20$~kN/m.
For details on the model parameters and structural properties, the reader is referred to \citet{ZhuPEM2023}.

In this application, we used the simulation software OpenSees \citep{mazzoni2006} to obtain the structural response under the specified ground motions.
The simulations of the structural response under a specified ground-motion time history are deterministic.
However, since multiple distinct signals can arise from a single ground motion parameter set $\bigl(I_a,\, t_{\mathrm{mid}},\,D_{5\text{--}95},\,\omega_g\bigr)$, the maximum interstory drift is a random variable with respect to these parameters.

In our study, different levels of fidelity are regulated by the time step of the simulations. 
The HF simulator uses a fine time step of $\delta t_\text{H} = 0.01~\text{s}$, whereas the LF simulator is derived from the same model by increasing, but with a seven times larger time step $\delta t_\text{L} = 0.07~\text{s}$.
The average wall‑clock runtime for an HF simulation is $\bar t_{\text{H}} = 10.3\,~\text{s}$, whereas the average runtime for an LF simulation is $\bar t_{\text{L}} = 1.8\,\text{s}$, meaning that an LF simulation is about 5.7 times faster than an HF simulation.

The process to generate the HF and LF data is the following: we first generate realizations of the random ground motion parameters using LHS and the corresponding ground motion signals; then we run a simulation for the three-story frame to obtain the maximum interstory drift. 
To keep the total number of simulations tractable, while being able to obtain all the HF and LF training data sets required for our $N_\text{r} = 25$ validation repetitions, we generate large HF and LF data pools and draw the required datasets by subsampling. 
To obtain the HF training data, we generate $10^4$ data points using the HF simulations, from which we randomly subsample $N_\text{H} \in \{100, 200, 400, 800\}$ HF data points.
For the LF training data, we generate $2 \times 10^4$ data using the LF simulations, from which we randomly subsample $N_\text{L} = 1,\!000 $ LF data. 

As mentioned above, the four input parameters are statistically dependent and coupled via a Gaussian copula (see \cref{tab:earthquake_inputvars}). 
However, for the PCEs used in constructing all GLaMs, the inputs are assumed to be independent, as detailed in \cref{sec:GLaM_single}. 
In such cases, it is common practice to disregard the input dependencies during model construction, as this often leads to more accurate pointwise predictions in data-driven settings compared to transforming the correlated inputs into uncorrelated ones via the highly nonlinear Rosenblatt transform \citep{Torre2019}. 
However, the input dependencies are still taken into account when generating training and validation samples.

\begin{figure}[htb]
    \centering
    \includegraphics[width=\textwidth,clip=true,trim=10 0 80 0]{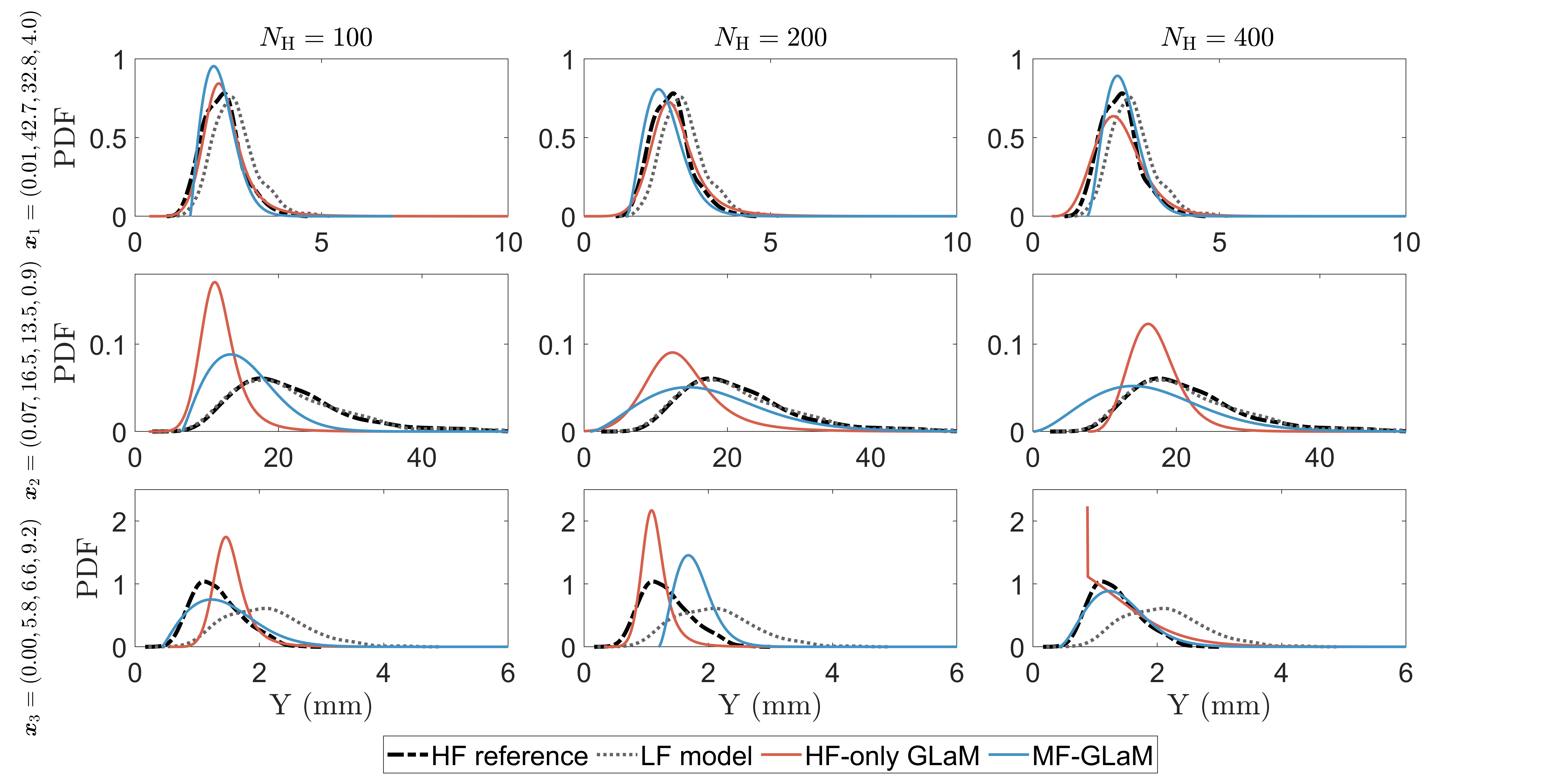}
    \caption{Multi-story building subject to an earthquake. Comparison of the true HF (black dashed curve) and LF (black dotted curve) conditional PDFs and the corresponding predictions from an MF-GLaM (blue curve) and a GLaM trained on the HF data only (red curve). The comparison is made for three random points and for increasing size of HF training data.}
    \label{fig:08Earthquake_PointApprox}
\end{figure}

\cref{fig:08Earthquake_PointApprox} compares the empirical response distributions from our true HF and LF simulators to the corresponding predicted distributions from the HF‐only GLaM (red curve) and the MF‐GLaM (blue curve).
Three random input samples obtained using LHS are shown in each row, and each column corresponds to a different HF training set size.
For each input sample, we generated $N_\text{R}=500$ HF runs and  $N_\text{R}=500$ LF runs to estimate the true HF (black dashed curve) and LF (gray dotted curve) distributions, respectively.
The output has been multiplied by $1,\!000$ for convenient display in millimeters. 

Despite the LF distribution often diverging from the corresponding HF response, the MF‐GLaM successfully captures the bulk and shape of the HF distribution across various scenarios, especially when HF data are scarce. 
In contrast, the HF‐only GLaM tends to be more sensitive to limited HF samples and can misalign with the empirical HF distribution, e.g., at $\ve{x}_2$ for $N_\text{H} =100$ or at $\ve{x}_3$ for $N_\text{H} =400$.

\begin{figure}[htb]
    \centering
    \includegraphics[width=0.6\textwidth]{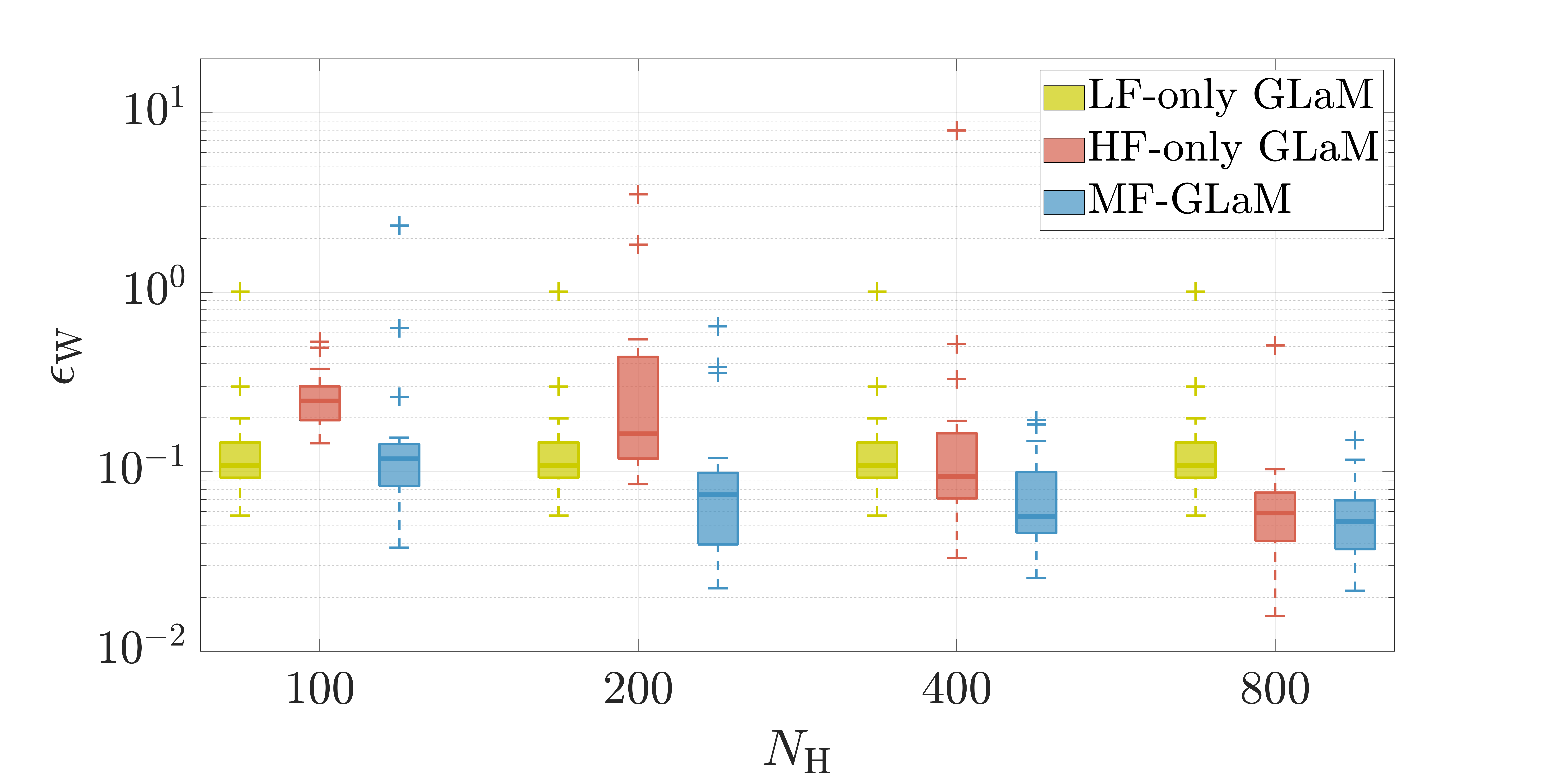}
    \caption{Multi-story building subject to an earthquake. Comparison of the normalized Wasserstein distance $\epsilon_\text{W}$ for LF-only, HF-only, and MF GLaMs across increasing HF experimental design sizes $N_\text{H}$.
    The box plots correspond to 25 repetitions of the procedure, and $\epsilon_\text{W}$ is computed using $200$ test points, with $250$ replications performed at each point.}
    \label{fig:08Earthquake_WassDist}
\end{figure}

\cref{fig:08Earthquake_WassDist} and the corresponding summary in \cref{tab:08Earthquake_WassDist_summary} present the normalized Wasserstein distance $\epsilon_{W}$ for three surrogate models, LF-only, HF-only, and MF-GLaM, across increasing high-fidelity training set sizes from $N_{\text{H}} = 100$ to $N_{\text{H}} = 800$. The reference validation data were generated by running our HF simulator on 200 Latin hypercube samples, each with $N_{\text{R}} = 250$ replications, resulting in a total of $50,\!000$ HF runs. 
In this application, the LF simulator is a numerically less refined version of the HF model, rather than a physically distinct one. This may explain why the LF-only GLaM performs relatively well at small HF training sizes ($N_{\text{H}} \le 200$), when the available HF data are too limited to fully capture the underlying variability.  
Overall, the MF-GLaM (blue boxes) consistently achieves the lowest error among the three emulators, with especially large performance gains at smaller HF sample sizes ($N_{\text{H}} \le 400$). 

Accounting for the computational cost of HF and LF runs, where 5.7 LF runs are considered equivalent in cost to one HF run, the total cost of using $N_H = 200$ HF samples and $N_L = 1,\!000$ LF samples corresponds to $200 + \frac{1,\!000}{5.7} \approx 375$ HF-equivalent runs. This configuration leads to an MF-GLaM with median $\epsilon_\text{W} =  0.075$ (see \cref{tab:08Earthquake_WassDist_summary}). This error is significantly lower than the median $\epsilon_\text{W}$ of the HF-only GLaM trained on $N_{\text{H}} = 400$ HF points, for which $\epsilon_\text{W} = 0.094$, indicating that the MF approach performs better at a reduced total cost. 
A similar trend is observed for the MF-GLaM trained on $N_H = 400$ and $N_L = 1,\!000$ points, where the total cost amounts to $400 + \frac{1,\!000}{5.7} \approx 575$ HF-equivalent runs.
In this case, the MF-GLaM's median error, $\epsilon_\text{W} = 0.056$, is comparable to that of the HF-only GLaM trained on $N_{\text{H}}=800$ HF points, for which $\epsilon_\text{W} = 0.059$.
Thus, comparable performance is achieved at a significantly lower computational cost, corresponding to a budget saving of approximately $28\%$.

\begin{table}[htb]
\centering
\caption{
Multi-story building subject to an earthquake. Median (interquartile range) of the normalized Wasserstein distance $\epsilon_\text{W}$ $\times10^{-2}$, computed over $N_\text{r}=25$ repetitions, comparing HF‑only GLaM and MF‑GLaM models for four HF training set sizes.
Baseline $\epsilon_\text{W}$, for the LF‑only GLaM: $10.8 \, (5.3) \times 10^{-2}$.
}
\label{tab:08Earthquake_WassDist_summary}
\begin{tabular}{c c c}
\toprule
$N_H$ & HF-only GLaM & MF-GLaM \\
\midrule
100  & 24.8 (10.6) & 11.8 (6.0) \\
200  & 16.3 (31.9) & 7.5 (5.9) \\
400  & 9.4 (9.3)   & 5.6 (5.4) \\
800  & 5.9 (3.5)   & 5.3 (3.2) \\
\bottomrule
\end{tabular}
\end{table}

\FloatBarrier

\section{Discussion and conclusions}
\label{sec:conclusions}
In this paper, we presented a multifidelity framework, namely multifidelity generalized lambda model (MF-GLaM), for efficiently emulating the conditional response distribution of a high-fidelity stochastic simulator by integrating information from both high-fidelity and low-fidelity stochastic simulators. 
Our approach extends the single-fidelity GLaM methodology \citep{ZhuIJUQ2020,ZhuSIAM2021}, in which the response distribution of a stochastic simulator is represented by a flexible four-parameter generalized lambda distribution. In our multifidelity setting, each GLD parameter of the HF model is decomposed into the corresponding LF parameter plus a discrepancy function, both expressed by polynomial chaos expansions. 
No repeated evaluations at the same input or control over internal random seeds are required:  
the coefficients of both components are estimated jointly by maximum likelihood from all the available HF and LF data.
The relative weight $p$ that balances the LF and HF contributions in the likelihood acts as a  regularizer; while in this study we adopt a neutral value $p=0.5$, it could, in principle, be treated as a tunable hyperparameter and learned from the data to balance the bias–variance trade-off.
We further derived a modified Bayesian information criterion to automatically select the polynomial bases of the discrepancy component.

Across three increasingly complex numerical examples, comprising two synthetic setups and one realistic earthquake application, the MF-GLaM consistently achieves faster convergence to the true HF conditional distributions compared to single-fidelity GLaM stochastic emulators trained only on HF or LF data. 
Pointwise comparisons of the conditional PDFs at different inputs show that MF-GLaM captures the bulk and shape of the HF distributions more accurately, especially in the presence of limited HF training data. 
Moreover, incorporating LF runs can substantially reduce the computational cost of emulating HF stochastic simulators. In the earthquake application, combining a modest HF training set with $1,\!000$ LF runs achieved performance comparable to training solely with a significantly larger HF dataset, resulting in approximately 28\% total cost savings. 
In  practical scenarios, the HF model can be significantly more computationally expensive than the LF model, often by multiple orders of magnitude (as opposed to the $5.7:1$ ratio in our example), which can make the efficiency gains offered by our framework even more substantial.
The framework is equally applicable when the LF and HF responses are defined on different scales, e.g.,  when the LF output is a correlated but distinct quantity of interest. 
In such cases, the data should be rescaled to a common scale before fitting the MF-GLaM, since the parameter-level fusion in MF-GLaM does not explicitly account for scale mismatches.

In future work, various extensions of the present framework can be explored.
First, adaptive cost-aware designs could be developed to determine the optimal HF-to-LF data ratio.
For instance, in our final application, comparable results might have been achievable with a smaller LF training dataset, potentially leading to even greater computational savings.
A second valuable direction would be to explore multiple LF models of varying fidelity (e.g., via different integration time steps in our final application) to find an optimal cost-accuracy tradeoff. 
In addition, while generalized lambda models suffice for representing most unimodal response distributions, extending MF-GLaM to handle multimodal distributions would further enhance the versatility of the framework, as an example by means of the recently introduced stochastic polynomial chaos expansions \citep{ZhuStoPCE2023}. 
Finally, exploring special cases, such as deterministic HF models with stochastic LF counterparts or vice versa, and LF models with higher input dimensionality than HF models, could be of practical interest.

\appendix

\section{Maximum likelihood estimator for the MF-GLaM}
\label{annex_A}

We aim to minimize the Kullback--Leibler (KL) divergence between the joint distributions $f_0$, given by \cref{eq:joint_model}, and $f_{\GLaM}$, given by \cref{eq:joint_model_glam}. 
This KL divergence is expressed as:
\begin{equation}\label{eq:KL2orig}
\begin{split}
    &\DL(f_0||f_{\GLaM}) = \int \log\left(\frac{f_0(s,\ve{x},y_H,y_L)}{f_{\GLaM}(s,\ve{x},y_H,y_L;\ve{\theta})}\right) f_0(s,\ve{x},y_H,y_L) \, \D s \, \D\ve{x} \, \D y_H \, \D y_L \\
    & = \int \log\left(\frac{p \mathbbm{1}_{\{s=L\}}f_{\ve{X}}(\ve{x}) f_{Y_L \mid \ve X}(y_L|\ve{x}) + (1-p)\mathbbm{1}_{\{s=H\}}f_{\ve{X}}(\ve{x}) f_{Y_H \mid \ve X}(y_H|\ve{x})}
    {p \mathbbm{1}_{\{s=L\}}f_{\ve{X}}(\ve{x}) f_{\GLaM}^\text L(y_L|\ve{x};\ve{c}) + (1-p)\mathbbm{1}_{\{s=H\}}f_{\ve{X}}(\ve{x}) f_{\GLaM}^\text{MF}(y_H|\ve{x}; \ve{c}, \ve{d})}\right) \\ 
    & \qquad\qquad \cdot f_0(s,\ve{x},y_H,y_L) \, \D s \, \D\ve{x} \, \D y_H \, \D y_L. 
\end{split}
\end{equation}
Assuming that the integrand is absolutely integrable, we apply Fubini's theorem to first integrate with respect to the counting measure $\mu_S$, as follows:
\begin{equation}
\begin{split}
    &\DL(f_0||f_{\GLaM}) = p \int \log\left(\frac{p f_{\ve{X}}(\ve{x}) f_{Y_L \mid \ve X}(y_L|\ve{x})}{p f_{\ve{X}}(\ve{x}) f_{\GLaM}^\text{L}(y_L|\ve{x};\ve{c})} \right) f_0(L,\ve{x},y_H, y_L)\, \D\ve{x} \, \D y_H \, \D y_L \\
    &\qquad\qquad + (1-p)\int \log\left(\frac{(1-p) f_{\ve{X}}(\ve{x}) f_{Y_H \mid \ve X}(y_H|\ve{x})}{(1-p) f_{\ve{X}}(\ve{x}) f_{\GLaM}^\text{MF}(y_H|\ve{x}; \ve{c}, \ve{d})} \right) f_0(H,\ve{x},y_H, y_L) \, \D\ve{x} \, \D y_H \, \D y_L \\
    & = p \int \log\left(\frac{f_{Y_L \mid \ve X}(y_L|\ve{x})}{f_{\GLaM}^\text{L}(y_L|\ve{x};\ve{c})} \right) f_{\ve{X}}(\ve{x})f_{Y_L \mid \ve X}(y_L\mid\ve{x}) \, \D\ve{x} \, \D y_H \, \D y_L \\
    &\qquad\qquad + (1-p)\int \log\left(\frac{f_{Y_H \mid \ve X}(y_H|\ve{x})}{f_{\GLaM}^\text{MF}(y_H|\ve{x}; \ve{c}, \ve{d})} \right) f_{\ve{X}}(\ve{x})f_{Y_H \mid \ve X}(y_H\mid\ve{x}) \, \D\ve{x} \, \D y_H \, \D y_L\\
    & = - p \int \log\left(f_{\GLaM}^\text{L}(y_L|\ve{x};\ve{c})\right) f_{\ve{X}}(\ve{x})f_{Y_L \mid \ve X}(y_L\mid\ve{x}) \, \D\ve{x} \, \D y_L \, + \, \text{const}_1 \\
    &\qquad\qquad - (1-p)\int \log\left(f_{\GLaM}^\text{MF}(y_H|\ve{x}; \ve{c}, \ve{d})\right) f_{\ve{X}}(\ve{x})f_{Y_H \mid \ve X}(y_H\mid\ve{x}) \, \D\ve{x} \, \D y_H \, + \, \text{const}_2\\
    & =  - p\Espe{\ve{X},Y_L|L}{\log\left(f_{\GLaM}^\text{L}(Y_L|\ve{X};\ve{c}) \right)} - (1-p)\Espe{\ve{X},Y_H|H}{\log\left(f_{\GLaM}^\text{MF}(Y_H|\ve{X}; \ve{c}, \ve{d}) \right)} +  \text{const}.
\end{split}
\end{equation}
Therefore, by minimizing the Kullback-Leibler divergence, we obtain the following parameters:
\begin{equation}
\label{eq:theta_1}
    \ve{\theta}^*=\arg\max_{\ve{\theta}} \: p\, \Espe{\ve{X},Y_L|L}{\log\left(f_{\GLaM}^\text{L}(Y_L|\ve{X};\ve{c}) \right)} + (1-p) \, \Espe{\ve{X},Y_H|H}{\log\left(f_{\GLaM}^\text{MF}(Y_H|\ve{X}; \ve{c}, \ve{d}) \right)},
\end{equation}
where $\ve{\theta} = (\ve{c}, \ve{d})$.

Although we could estimate the parameters directly from \cref{eq:theta_1} by replacing the two expectations with empirical means from the HF and LF samples, we instead aim to derive a log-likelihood estimator for $\ve{\theta}$ in a way to to associate each sample an additional weight, in order to ensure consistency with the Bayesian derivation of the BIC, which we will present later. The weights are obtained via \emph{importance sampling}.
When generating samples within a multifidelity setting, we would typically not sample $S$ following the Bernoulli distribution with $p$ but with a much higher probability $p'$ towards the low-fidelity model. 
For a design of $N_\text L$ model evaluations for the LF model and $N_\text H$ for the HF model, the sampling probability can be regarded as $p' = \frac{N_\text L}{N_\text L+N_\text H}$. Therefore, we consider an instrumental distribution
\begin{equation}
    f_1\bigl(s,\ve{x},y_H,y_L\bigr) 
    \;=\;
    p' \mathbbm{1}_{\{s=L\}} f_{\ve{X}}(\ve{x}) f_{Y_L \mid \ve X}(y_L|\ve{x}) 
    \;+\;
    (1-p')\mathbbm{1}_{\{s=H\}} f_{\ve{X}}(\ve{x}) f_{Y_H \mid \ve X}(y_H|\ve{x}).
\end{equation}



This allows us to re-write the KL divergence calculation through importance sampling, as follows:
\begin{equation}    
\begin{split}
    \DL(f_0||f_{\GLaM}) &= \int \log\left(\frac{f_0(s,\ve{x},y_H,y_L)}{f_{\GLaM}(s,\ve{x},y_H,y_L;\ve{\theta})}\right) 
    \frac{f_0(s,\ve{x},y_H,y_L)}{f_1(s,\ve{x},y_H,y_L)}  \cdot f_1(s,\ve{x},y_H,y_L) 
    \, \D s \, \D\ve{x} \, \D y_H \, \D y_L.
\end{split}    
\end{equation}

Similar to the previous derivation, this quantity can be computed as
\begin{equation}\label{eq:IS}
\begin{split}
    \DL&(f_0||f_{\GLaM}) = 
    \int \log\left(\frac{f_0(L,\ve{x},y_H,y_L)}{f_{\GLaM}(L,\ve{x},y_H,y_L;\ve{\theta})}\right) \frac{f_0(L,\ve{x},y_H,y_L)}{f_1(L,\ve{x},y_H,y_L)} f_1(L,\ve{x},y_H,y_L) \, \D\ve{x} \, \D y_H \, \D y_L \\ 
    &+ \int \log\left(\frac{f_0(H,\ve{x},y_H,y_L)}{f_{\GLaM}(H,\ve{x},y_H,y_L;\ve{\theta})}\right) \frac{f_0(H,\ve{x},y_H,y_L)}{f_1(H,\ve{x},y_H,y_L)} f_1(H,\ve{x},y_H,y_L) \, \D\ve{x} \, \D y_H \, \D y_L.
\end{split}
\end{equation}
The importance factor for low-fidelity samples is:
\begin{equation}\label{eq:ISweight_LF}
    w_L = \frac{f_0\bigl(L,\ve{x},y_H,y_L\bigr)}{f_1\bigl(L,\ve{x},y_H,y_L\bigr)} = \frac{p}{p'},
\end{equation}
while the importance factor for high-fidelity samples is:
\begin{equation}\label{eq:ISweight_HF}
    w_H = \frac{f_0\bigl(H,\ve{x},y_H,y_L\bigr)}{f_1\bigl(H,\ve{x},y_H,y_L\bigr)} = \frac{(1-p)}{(1-p')}.
\end{equation}
Therefore, each LF and HF sample is weighed by 
\begin{equation}\label{eq:ISweight}
    w_L = \frac{p\,(N_\text L+N_\text H)}{N_\text L} \;
    \text{ and }\;
    w_H =\frac{(1-p)(N_\text L+N_\text H)}{N_\text H},
\end{equation}
respectively.

By sampling through the instrumental distribution, we obtain the weighted log-likelihood of the joint distribution of $(S,\ve{X},Y_H, Y_L)$ that we need to maximize, which is given by

\begin{equation}
\begin{split}
&\ell(\ve{\theta};\cs, \ve \cx_\text H, \ve \cx_\text L,\cy_\text H, \cy_\text L) = \sum_{i=1}^{N_\text{L}} w_L\log\left(f_{\GLaM}^\text{L}(y^{(i)}_\text{L}|\ve{x}_\text{L}^{(i)};\ve{c})\right) + \sum_{i=1}^{N_\text{H}} w_H\log\left(f_{\GLaM}^\text{MF}(y^{(i)}_\text{H}|\ve{x}_\text{H}^{(i)}; \ve{c}, \ve{d})\right) + \text{const}\\
&= \frac{p(N_\text{L}+N_\text{H})}{N_\text{L}}\sum_{i=1}^{N_\text{L}} \log\left(f_{\GLaM}^\text{L}(y^{(i)}_\text{L}|\ve{x}_\text{L}^{(i)};\ve{c})\right) + \frac{(1-p)(N_\text{L}+N_\text{H})}{N_\text{H}}\sum_{i=1}^{N_\text{H}} \log\left(f_{\GLaM}^\text{MF}(y^{(i)}_\text{H}|\ve{x}_\text{H}^{(i)}; \ve{c}, \ve{d})\right) + \text{const}.
\label{eq:llh_annexA}
\end{split}
\end{equation}

Therefore, by maximizing $\ell$ , we obtain the following parameters:
\begin{equation}
    \ve{\theta}^*=\arg\max_{\ve{\theta}} \: \frac{p(N_\text{L}+N_\text{H})}{N_\text{L}}\sum_{i=1}^{N_\text{L}} \log\left(f_{\GLaM}^\text{L}(y^{(i)}_\text{L}|\ve{x}_\text{L}^{(i)};\ve{c})\right) + \frac{(1-p)(N_\text{L}+N_\text{H})}{N_\text{H}}\sum_{i=1}^{N_\text{H}} \log\left(f_{\GLaM}^\text{MF}(y^{(i)}_\text{H}|\ve{x}_\text{H}^{(i)}; \ve{c}, \ve{d})\right).
\end{equation}

\section{Bayesian information criterion for MF-GLaMs}
\label{annex_B}

\label{subsubsec:BICderiv}
In this section, we provide a derivation for the Bayesian information criterion to choose the most suitable model among different multifidelity generalized lambda models. 
The derivation closely follows the steps from \citet{Neath2012}, who originally derived the BIC in a general single-fidelity setting.

Let us denote as $\ve D$ the random vector $(S, \ve X, Y_\text H, Y_\text L)$ and $\ve \cd = (\cs, \ve \cx_\text H, \ve \cx_\text L,\cy_\text H, \cy_\text L)$ the entire collection of the observed HF and LF data.
As a reminder, we denote the full set of parameters to be estimated as $\ve \theta = (\ve c, \ve d)$. As seen from \cref{eq:llh}, the parameters $\ve c, \ve d$ are estimated jointly from the HF and LF data, with $\ve c$ depending on both HF and LF data, while $\ve d$ depending only on the HF data.

Our goal is to select the model that best describes the observed data $\ve \cd$ from a set of candidate models $\cm_k, \, k=1,..., N_\text m$. 
In the following, we denote the number of parameters in the full set $\ve \theta_k$ of the model $\cm_k$ by $n_{\ve \theta_k} \coloneqq \text{card}(\ve \theta_k)$, and the number of parameters in $\ve c_k$ by $n_{\ve c_k} \coloneqq \text{card}(\ve c_k)$.

We define $\pi(\cm_k)$ as a discrete prior over the models $\cm_k$, and denote $g(\ve\theta_k \mid \cm_k)$ the prior distribution of parameters $\ve\theta_k$, given model $\cm_k$. The joint posterior of $\cm_k$ and $\ve\theta_k$ can be obtained from  Bayes' theorem as
\begin{align}
\Pro((\cm_k,\ve\theta_k) \mid \ve \cd) &= \frac{\pi(\cm_k) g(\ve\theta_k \mid \cm_k) \Pro(\ve \cd \mid \ve\theta_k)}{f_{\ve D}(\ve \cd)} \\
&= \frac{\pi(\cm_k) g(\ve\theta_k \mid \cm_k) L(\ve\theta_k \mid \ve \cd)}{f_{\ve D}(\ve \cd)},
\end{align}
where $f_{\ve D}(\ve \cd)$ denotes the marginal distribution of $\ve \cd$ and $L(\ve\theta_k \mid \ve \cd)$ is the likelihood of the parameters for the model $\cm_k$ given the data, such that $\log{L(\ve\theta_k \mid \ve \cd)} = \ell(\ve\theta_k \mid \ve \cd)$ in \cref{eq:llh}:
\begin{equation}
\begin{split}
\label{eq:llh_annexB}
\ell(\ve{\theta}_k;\cd) = 
\frac{(N_\text{L}+N_\text{H})}{2 N_\text{L}}\sum_{i=1}^{N_\text{L}} \log\left(f_{\GLaM}^\text{L}(y^{(i)}_\text{L}|\ve{x}_\text{L}^{(i)};\ve{c}_k)\right) 
+ \frac{(N_\text{L}+N_\text{H})}{2N_\text{H}}\sum_{i=1}^{N_\text{H}} \log\left(f_{\GLaM}^\text{MF}(y^{(i)}_\text{H}|\ve{x}_\text{H}^{(i)}; \ve{c}_k, \ve{d}_k)\right),
\end{split}
\end{equation}
which arises from \cref{eq:llh_annexA} with $p=0.5$, as discussed in \cref{sec:estimation}.
Then, the posterior probability for model $\cm_k$ is
\begin{equation}
\Pro(\cm_k \mid \ve \cd) = f_{\ve D}(\ve \cd)^{-1} \pi(\cm_k) \int L(\ve \theta_k \mid \ve \cd) g(\ve \theta_k \mid \cm_k) \, d\ve \theta_k.
\end{equation}
To select the model that best describes the observed data, we aim at maximizing this probability with respect to $k$, or equivalently, minimizing $-\log \Pro(\cm_k \mid \ve \cd)$. The latter reads:
\begin{equation}
-\log \Pro(\cm_k \mid \ve \cd) = \log  f_{\ve D}(\ve \cd)  - \log \pi(\cm_k) - \log \left\{ \int L(\ve \theta_k \mid \ve \cd) g(\ve \theta_k \mid \cm_k) \, d\ve \theta_k \right\}.
\end{equation}

Since $\log f_{\ve D}(\ve \cd)$ is independent of $k$, our goal is to minimize
\begin{equation} \label{eq:IC}
\text{IC}(\cm_k \mid \ve \cd) = - \log \pi(\cm_k) - \log \left\{ \int L(\ve \theta_k \mid \ve \cd) g(\ve \theta_k \mid \cm_k) \, d\ve \theta_k \right\}.
\end{equation}

To approximate the integrand in \cref{eq:IC}, we expand the log-likelihood about the vector of estimates ${\ve \theta}^*_k$ obtained by maximizing $\log L(\ve \theta_k \mid \ve \cd)$. Using second-order Taylor series expansion, we have:
\begin{align}
\log L(\ve \theta_k \mid \ve \cd) \equiv \ell(\ve\theta_k \mid \ve \cd)
&\approx \ell(\ve \theta^*_k \mid \ve \cd) + (\ve \theta_k - \ve \theta^*_k)^\top \frac{\partial \ell(\ve \theta_k \mid \ve \cd)}{\partial \ve \theta_k}\bigg|_{\ve \theta_k=\ve \theta^*_k} \nonumber\\
&+ \tfrac{1}{2}(\ve \theta_k - \ve \theta^*_k)^\top \frac{\partial^2 \ell(\ve \theta_k \mid \ve \cd)}{\partial \ve \theta_k \partial \ve \theta_k^\top}\bigg|_{\ve \theta_k=\ve \theta^*_k} (\ve \theta_k - \ve \theta^*_k).
\end{align}

Because $\ve \theta^*_k$ is the maximizer of $\ell(\ve \theta_k \mid \ve \cd)$, the first derivative at $\ve \theta^*_k$ vanishes. Thus,
\begin{align}
\ell(\ve \theta_k \mid \ve \cd) &\approx \ell(\ve \theta^*_k \mid \ve \cd) + \tfrac{1}{2}(\ve \theta_k - \ve \theta^*_k)^\top \frac{\partial^2 \ell(\ve \theta_k \mid \ve \cd)}{\partial \ve \theta_k \partial \ve \theta_k^\top}\bigg|_{\ve \theta_k=\ve \theta^*_k} (\ve \theta_k - \ve \theta^*_k) \\
&= \ell(\ve \theta^*_k \mid \ve \cd) - \tfrac{1}{2}\,
(\ve{\theta}_k - \ve{\theta}_k^{*})^{\mathsf{T}}
\;\mathbf{H}\;
(\ve{\theta}_k - \ve{\theta}_k^{*}),  \label{eq:ll_aprox}
\end{align}
where  
\begin{equation}\label{eq:Hessian}
\mathbf{H} = -\frac{\partial^2 \ell(\ve\theta_k \mid \ve \cd)}{\partial \ve\theta_k \partial \ve\theta_k} \bigg|_{\ve \theta_k = \ve{\theta}_k^{*}} 
\end{equation}
is the Hessian of the negative log-likelihood evaluated at $\ve{\theta}_k^{*}$.
The matrix $\mathbf{H}$ is of size $n_{\ve \theta_k} \times n_{\ve \theta_k}$, and each of its elements is defined as ${H}_{mn} = -\frac{\partial^2 \ell(\ve\theta_k \mid \ve \cd)}{\partial \theta_{k,m} \partial \theta_{k,n}} \bigg|_{\ve \theta_k = \ve \theta_k^{*}}$.

If we substitute the log-likelihood from \cref{eq:llh} into \cref{eq:Hessian}, the Hessian matrix reads:
\begin{align}
    \mathbf{H} &= -\frac{\partial^2 \ell(\ve\theta_k \mid \ve \cd)}{\partial \ve\theta_k \partial \ve\theta_k} \bigg|_{\ve \theta_k = \ve{\theta}_k^{*}} \\
    &=  \Bigg[-\frac{\partial^2 }{\partial \ve\theta_k \partial \ve\theta_k} \left ( \frac{N_\text H + N_\text L}{2} \frac{1}{N_\text L} \sum_{i=1}^{N_\text L} \log\left(f_{\GLaM}^\text{L}(y^{(i)}_\text{L}|\ve{x}_\text{L}^{(i)};\ve{c})\right) \right )  \nonumber\\
    & \qquad - \frac{\partial^2 }{\partial \ve \theta_k \partial \ve \theta_k} \left ( \frac{N_\text H + N_\text L}{2} \frac{1}{N_\text H} \sum_{i=1}^{N_\text H} \log\left(f_{\GLaM}^\text{MF}(y^{(i)}_\text{H}|\ve{x}_\text{H}^{(i)}; \ve{c}, \ve{d})\right) \right ) \Bigg]_{\ve \theta_k = \ve \theta_k^{*}} \\
    &= -\frac{N_\text H + N_\text L}{2} \frac{1}{N_\text L}\frac{\partial^2 \sum_{i=1}^{N_\text L}  \log L(\ve c_k \mid y_\text{L}^{(i)}, \ve{x}_\text{L}^{(i)})}{\partial \ve \theta_k \partial \ve \theta_k}  \Bigg|_{\ve \theta_k = \ve{\theta}_k^{*}} \nonumber \\
    & \qquad - \frac{N_\text H + N_\text L}{2} \frac{1}{N_\text H}\frac{\partial^2 \sum_{i=1}^{N_\text H}  \log L(\ve \theta_k \mid y_\text{H}^{(i)}, \ve{x}_\text{H}^{(i)})}{\partial \ve \theta_k \partial \ve \theta_k}  \Bigg|_{\ve \theta_k = \ve{\theta}_k^{*}},
\end{align}
where $L(\ve c_k \mid y_\text{L}^{(i)}, \ve{x}_\text{L}^{(i)}) = f_{\GLaM}^\text{L}(y^{(i)}_\text{L}|\ve{x}_\text{L}^{(i)};\ve{c})$ is the likelihood of the LF model parameters $\ve{c}_k$, given the $i^\text{th}$ LF observation , and similarly, $L(\ve \theta_k \mid y_\text{H}^{(i)}, \ve{x}_\text{H}^{(i)}) = f_{\GLaM}^\text{MF}(y^{(i)}_\text{H}|\ve{x}_\text{H}^{(i)}; \ve{c}, \ve{d})$ is the  likelihood of the of the MF model parameters $\ve{c}_k, \ve{d}_k$, given the $i^\text{th}$ HF observation.

Then, 
\begin{equation} \label{eq:Hessian_fisher_sum}
    \mathbf{H} = \frac{N_\text H + N_\text L}{2} (\bar I_{\text L} + \bar I_{\text H}),
\end{equation}
where
\begin{equation}
    \bar I_{\text H} = \bar I_{\text H}(\ve{\theta}_k^{*}, \ve \cd_\text H) = 
    -\frac{1}{N_\text H}\frac{\partial^2 \sum_{i=1}^{N_\text H}  \log L(\ve \theta_k \mid y_H^{(i)})}{\partial \ve \theta_k \partial \ve \theta_k} \bigg|_{\ve \theta_k = \ve{\theta}_k^{*}}    
\end{equation}
is the $n_{\ve \theta_k} \times n_{\ve \theta_k}$ average observed Fisher information matrix for the HF data, and 
\begin{equation}
    \bar I_{\text L} = \bar I_{\text L}(\ve{\theta}_k^{*}, \ve \cd_\text L) = \bar I_{\text L}(\ve c_k^{*}, \ve \cd_\text L) = 
    -\frac{1}{N_\text L}\frac{\partial^2 \sum_{i=1}^{N_\text L}  \log L(\ve c_k \mid y_L^{(i)})}{\partial \ve \theta_k \partial \ve \theta_k} \bigg|_{\ve \theta_k = \ve{\theta}_k^{*}}    
\end{equation} 
is the $n_{\ve \theta_k} \times n_{\ve \theta_k}$ average observed Fisher information matrix for the LF data, where the elements $\bar I_{\text L, ij} = 0, \, \forall i,j > n_{\ve c_k}$.
In the above, we have denoted $\ve\cd_\text{H} = (\ve \cx_\text{H}, \cy_\text{H})$ and $\ve\cd_\text{L} = (\ve \cx_\text{L}, \cy_\text{L})$.

Let $\bar I$ be the average observed Fisher information matrix for the full set of parameters computed on all the HF and LF data, such that 
\begin{equation}
    \bar I = \bar I(\ve{\theta}_k^{*}, \ve \cd) = \bar I_{\text H} + \bar I_{\text L}.
\end{equation}
From \cref{eq:Hessian_fisher_sum} we have:
\begin{equation}\label{eq:Hessian_Fisher}
    \mathbf{H} = \frac{N_\text H + N_\text L}{2} \bar I .
\end{equation}

Substituting $\mathbf{H}$ back into the log-likelihood expansion in \cref{eq:ll_aprox} gives:
\begin{equation}
    \ell(\ve \theta_k \mid \ve \cd) \equiv \log L(\ve \theta_k \mid \ve \cd) \approx \ell(\ve \theta^*_k \mid \ve \cd) - \tfrac{1}{2}(\ve \theta_k - \ve \theta^*_k)^\top \left [ \frac{N_\text H + N_\text L}{2} \bar I  \right] (\ve \theta_k - \ve \theta^*_k).
\end{equation}
Thus,
\begin{equation}
    L(\ve \theta_k \mid \ve \cd) \approx L(\ve \theta^*_k \mid \ve \cd) \cdot \exp\left\{-\tfrac{1}{2}(\ve \theta_k - \ve \theta^*_k)^\top \left [ \frac{N_\text H + N_\text L}{2} \bar I  \right] (\ve \theta_k - \ve \theta^*_k)\right\},
\end{equation}

Then, the information criterion in \cref{eq:IC} becomes:
\begin{align}
    &\text{IC}(\cm_k \mid \ve \cd) = - \log \pi(\cm_k) - \log \left\{ \int L(\ve \theta_k \mid \ve \cd) g(\ve \theta_k \mid \cm_k) \, d\ve \theta_k \right\} \nonumber\\
    &\approx - \log \pi(\cm_k) -  L(\ve \theta^*_k \mid \ve \cd) \int \exp\left\{-\tfrac{1}{2}(\ve \theta_k - \ve \theta^*_k)^\top \left [ \frac{N_\text H + N_\text L}{2} \bar I  \right] (\ve \theta_k - \ve \theta^*_k)\right\} g(\ve \theta_k \mid \cm_k) \, d\ve \theta_k 
\end{align}

If we consider $g(\ve \theta_k \mid \cm_k) = 1$ for an uninformative, flat prior, then we have:
\begin{equation} \label{eq:IC_with_integral}
    \text{IC}(\cm_k \mid \ve \cd) \approx - \log \pi(\cm_k) -  L(\ve \theta^*_k \mid \ve \cd) \int \exp\left\{-\tfrac{1}{2}(\ve \theta_k - \ve \theta^*_k)^\top \left [ \frac{N_\text H + N_\text L}{2} \bar I  \right] (\ve \theta_k - \ve \theta^*_k)\right\} \, d\ve \theta_k
\end{equation}



We notice that the integrand of the integral on the right side of \cref{eq:IC_with_integral} can be directly compared with the multivariate Gaussian PDF.
Thus, the integral equals:
\begin{align}
    \int \exp\left\{-\tfrac{1}{2}(\ve \theta_k - \ve \theta^*_k)^\top \left [ \frac{N_\text H + N_\text L}{2} \bar I  \right] (\ve \theta_k - \ve \theta^*_k)\right\}\, d\ve \theta_k &= 
    (2\pi)^{\frac{n_{\ve \theta_k}}{2}} \left | \frac{N_\text H + N_\text L}{2} \bar I \right |^{-\tfrac{1}{2}} \nonumber\\ 
    &= (2\pi)^{\frac{n_{\ve \theta_k}}{2}} \left ( \frac{N_\text H + N_\text L}{2} \right )^{-\frac{n_{\ve \theta_k}}{2}} \left | \bar I \right |^{-\tfrac{1}{2}}
\end{align}


Substituting this back into the right-hand side of \cref{eq:IC_with_integral} gives:
\begin{align}
&\text{IC}(\cm_k \mid \ve \cd) \approx - \log \{\pi(\cm_k)\} -  \log \left\{ L(\ve \theta^*_k \mid \ve \cd) (2\pi)^{\frac{n_{\ve \theta_k}}{2}} \left ( \frac{N_\text H + N_\text L}{2} \right )^{-\frac{n_{\ve \theta_k}}{2}} \left | \bar I \right |^{-\frac{1}{2}} \right\} \\
&= - \log \{\pi(\cm_k)\} -  \log L(\ve \theta^*_k \mid \ve \cd) - \frac{1}{2}n_{\ve \theta_k}\log (2\pi) 
+  \frac{1}{2} n_{\ve \theta_k} \log \left ( \frac{N_\text H + N_\text L}{2} \right ) + \frac{1}{2} \log \left | \bar I \right |.
\end{align}

For large amounts of HF and LF data, $N_\text H$ and $N_\text L$ respectively, terms that do not grow with them become negligible for model selection. Ignoring those terms, the information criterion we have to minimize reads:
\begin{equation} \label{eq:bic_final1}
\text{IC}(\cm_k \mid \ve \cd) \approx - \log L(\ve \theta^*_k \mid \ve \cd) + \frac{1}{2} n_{\ve \theta_k} \log \left ( \frac{N_\text H + N_\text L}{2} \right ).
\end{equation}
The right-hand side of \cref{eq:bic_final1} is the BIC estimate for an MF-GLaM \( \cm_k \), which can equivalently be written as:
\begin{equation} 
\text{MF-BIC}(\cm_k \mid \ve \cd) = - 2 \log L(\ve \theta^*_k \mid \ve \cd) + n_{\ve \theta_k} \log \left ( \frac{N_\text H + N_\text L}{2} \right ).
\end{equation}



\section*{Declaration of Competing Interest}
The authors declare that they have no known competing financial interests or personal relationships that could have appeared to influence the work reported in this paper.

\section*{Acknowledgments}
This research was supported by the GREYDIENT project, funded by the EU Horizon 2020 program under the Marie Sk\l{}odowska-Curie grant agreement No. 955393, whose support is gratefully acknowledged.

\bibliography{bibliography.bib}

\end{document}